\documentclass{article}

     \PassOptionsToPackage{numbers, compress}{natbib}

 \usepackage[preprint]{neurips_2019}

\usepackage[utf8]{inputenc} 
\usepackage[T1]{fontenc}    
\usepackage{hyperref}       
\usepackage{url}            
\usepackage{booktabs}       
\usepackage{amsfonts}       
\usepackage{nicefrac}       
\usepackage{microtype}      

\usepackage{graphics}      
\usepackage{savesym}
\usepackage{amsmath}
\usepackage{color}
\usepackage{ccicons}          
\usepackage{relsize} 
\usepackage{mathtools,xparse}
\usepackage[table]{xcolor}
\usepackage{verbatim}
\usepackage{comment}
\usepackage{soul}
\usepackage{multicol} 
\usepackage{multirow}
\usepackage{colortbl} 

\usepackage{amssymb}
\usepackage{wasysym}
\makeatletter
\def\oversortoftilde#1{\mathop{\vbox{\m@th\ialign{##\crcr\noalign{\kern3\p@}%
				\sortoftildefill\crcr\noalign{\kern3\p@\nointerlineskip}%
				$\hfil\displaystyle{#1}\hfil$\crcr}}}\limits}

\def\sortoftildefill{$\m@th \setbox\z@\hbox{$\braceld$}%
	\braceld\leaders\vrule \@height\ht\z@ \@depth\z@\hfill\braceru$}
\makeatother
\usepackage{tikz-cd}
\usepackage{tabulary}

\usepackage[english]{babel}
\usepackage{enumitem}

\newtheorem{theorem}{Theorem}

\newtheorem{lemma}{Lemma}

\makeatletter
\renewcommand*{\@opargbegintheorem}[3]{\trivlist
	\item[\hskip \labelsep{\bfseries #1\ #2}] \textbf{(#3)}\ \itshape}
\makeatother

\usepackage{array}
\usepackage{bytefield}

\usepackage[labelformat=simple,font={normal}, labelsep=period,singlelinecheck=on]{caption}
\usepackage[labelformat=simple,font={normal}, labelsep=period,singlelinecheck=on]{subcaption}

\usepackage{cleveref}
\crefname{section}{§}{§§}
\Crefname{section}{§}{§§}

\usepackage{slashbox}
\usepackage{multirow}
\usepackage{xcolor}

\usepackage{dsfont}
\usepackage{transparent}
\usepackage{import}
\usepackage{calrsfs}
\DeclareMathAlphabet{\pazocal}{OMS}{zplm}{m}{n}
\newcommand{\La}{\pazocal{L}}

\newcommand{\Ta}{\pazocal{T}}

\newcommand{\Fa}{\pazocal{F}}
\newcommand{\Ra}{\pazocal{R}}
\newcommand{\Na}{\pazocal{N}}

\newcommand{\Ka}{\pazocal{K}}
\newcommand{\BigO}{\pazocal{O}}

\DeclareMathAlphabet\mathbfcal{OMS}{cmsy}{b}{n}

\usepackage{amsfonts}

\newcolumntype{C}{c<{\kern\tabcolsep}@{}}

\usepackage{tikz}
\definecolor{mygreen}{RGB}{0,176,80}
\definecolor{myred}{RGB}{102,0,0}
\definecolor{myblue}{RGB}{0,0,102}
\definecolor{myblue2}{RGB}{0,51,102}
\definecolor{myblue3}{RGB}{0,76,153}   
\definecolor{myblue4}{RGB}{48, 144, 199}  

\usetikzlibrary{decorations.pathreplacing}
\usetikzlibrary{bending}
\usepackage{lipsum}

\usetikzlibrary{positioning}
\tikzset{main node/.style={circle,fill=blue!20,draw,minimum size=1cm,inner sep=0pt},
}

\usepackage{etoolbox}
\patchcmd{\footnotemark}{\stepcounter{footnote}}{\refstepcounter{footnote}}{}{}

\usepackage{tabulary}
\newcolumntype{K}[1]{>{\centering\arraybackslash}p{#1}}
\usepackage{makecell}

\usepackage[leftcaption]{sidecap}
\sidecaptionvpos{figure}{t}
\usepackage{wrapfig}
\usepackage{picins}

\usepackage{amsmath,upgreek,bm}

\usepackage{tikzit}
\usepackage{pspicture}
\usepackage{pstricks}
\usepackage{graphicx}
\usepackage{amssymb}
\usepackage{wasysym}
\usetikzlibrary{positioning}
\usetikzlibrary{shapes,arrows}
\usetikzlibrary{calc}
\usetikzlibrary{decorations.pathreplacing}

\makeatletter
\pgfkeys{/pgf/.cd,
	parallelepiped offset x/.initial=5mm,
	parallelepiped offset y/.initial=5mm
}
\pgfdeclareshape{parallelepiped}
{
	\inheritsavedanchors[from=rectangle] 
	\inheritanchorborder[from=rectangle]
	\inheritanchor[from=rectangle]{north}
	\inheritanchor[from=rectangle]{north west}
	\inheritanchor[from=rectangle]{north east}
	\inheritanchor[from=rectangle]{center}
	\inheritanchor[from=rectangle]{west}
	\inheritanchor[from=rectangle]{east}
	\inheritanchor[from=rectangle]{mid}
	\inheritanchor[from=rectangle]{mid west}
	\inheritanchor[from=rectangle]{mid east}
	\inheritanchor[from=rectangle]{base}
	\inheritanchor[from=rectangle]{base west}
	\inheritanchor[from=rectangle]{base east}
	\inheritanchor[from=rectangle]{south}
	\inheritanchor[from=rectangle]{south west}
	\inheritanchor[from=rectangle]{south east}
	\backgroundpath{
		\southwest \pgf@xa=\pgf@x \pgf@ya=\pgf@y
		\northeast \pgf@xb=\pgf@x \pgf@yb=\pgf@y
		\pgfmathsetlength\pgfutil@tempdima{\pgfkeysvalueof{/pgf/parallelepiped offset x}}
		\pgfmathsetlength\pgfutil@tempdimb{\pgfkeysvalueof{/pgf/parallelepiped offset y}}
		\def\ppd@offset{\pgfpoint{\pgfutil@tempdima}{\pgfutil@tempdimb}}
		\pgfpathmoveto{\pgfqpoint{\pgf@xa}{\pgf@ya}}
		\pgfpathlineto{\pgfqpoint{\pgf@xb}{\pgf@ya}}
		\pgfpathlineto{\pgfqpoint{\pgf@xb}{\pgf@yb}}
		\pgfpathlineto{\pgfqpoint{\pgf@xa}{\pgf@yb}}
		\pgfpathclose
		\pgfpathmoveto{\pgfqpoint{\pgf@xb}{\pgf@ya}}
		\pgfpathlineto{\pgfpointadd{\pgfpoint{\pgf@xb}{\pgf@ya}}{\ppd@offset}}
		\pgfpathlineto{\pgfpointadd{\pgfpoint{\pgf@xb}{\pgf@yb}}{\ppd@offset}}
		\pgfpathlineto{\pgfpointadd{\pgfpoint{\pgf@xa}{\pgf@yb}}{\ppd@offset}}
		\pgfpathlineto{\pgfqpoint{\pgf@xa}{\pgf@yb}}
		\pgfpathmoveto{\pgfqpoint{\pgf@xb}{\pgf@yb}}
		\pgfpathlineto{\pgfpointadd{\pgfpoint{\pgf@xb}{\pgf@yb}}{\ppd@offset}}
	}
}
\makeatother

\newsavebox\CBox

\usepackage[capbesideposition=outside,capbesidesep=quad]{floatrow}

\title{Learning Universal Graph Neural Network Embeddings With Aid Of Transfer Learning}

\author{%
  Saurabh Verma \\
  Department of Computer Science\\
  University of Minnesota Twin Cities\\
  \texttt{verma076@cs.umn.edu} \\
   \And
   Zhi-Li Zhang \\
  Department of Computer Science\\
University of Minnesota Twin Cities\\
\texttt{zhang@cs.umn.edu} \\
}

\begin{document}

\maketitle

\begin{abstract}

Learning powerful data embeddings has become a center piece in machine learning,  especially in natural language processing  and computer vision  domains. The crux of these embeddings is that they are pretrained  on huge corpus of data in a unsupervised fashion, sometimes aided with transfer learning. However currently in the graph learning domain, embeddings learned through existing graph  neural networks (GNNs) are  task dependent and  thus cannot be shared across different datasets. In this paper, we present a  first powerful and theoretically guaranteed graph neural network that is designed to learn {\em task-independent} graph embeddings,  thereafter referred to as {\em deep  universal graph embedding} (\textsc{DUGnn}). Our \textsc{DUGnn}  model 
incorporates a novel graph neural network (as a universal graph encoder) and leverages rich Graph Kernels (as a multi-task graph decoder) for both unsupervised learning and (task-specific) adaptive supervised learning. 
By learning task-independent graph embeddings across diverse datasets, \textsc{DUGnn} also reaps the benefits of transfer learning. 
Through extensive experiments and ablation studies, we show that the proposed \textsc{DUGnn}  model consistently outperforms both the existing state-of-art GNN models and Graph Kernels by an increased accuracy of $\mathbf{3\textbf{\%}-8\textbf{\%}}$ on graph classification benchmark datasets.

\end{abstract}

\section{Introduction}

Learning powerful data embeddings   has become a center piece in machine learning for producing superior results. This new trend of learning embeddings from data can be attributed to the  huge success  of word2vec~\cite{mikolov2013distributed, pennington2014glove}  with unprecedented  real-world performance in natural language processing (NLP). The importance of  extracting high quality  embeddings has now been  realized   in many other domains such as computer vision (CV)  and recommendation systems~\cite{kiela2014learning, karpathy2014deep, zhang2016collaborative}. The crux of these embeddings is that they are  pretrained  in an unsupervised fashion on huge amount of data  and thus can potentially capture all kind of contextual information. Such embedding learning is further advanced by the incorporation of multi-task learning and transfer learning which allow more generalized embeddings to be learned across different datasets. These developments have brought major breakthroughs in NLP: DecaNLP~\cite{mccann2018natural} and BERT~\cite{devlin2018bert} are  recent such prime examples. 

Besides natural languages (``sequence'' or 1D data) and images (2D or 3D) which are well-structured, the idea of embedding learning has been applied to (irregular) ``graph-structured'' data for various graph learning tasks, such as node classification or link prediction. Unlike word embeddings where vocabulary is typically finite, graph ``vocabulary'' is potentially infinite (i.e, count of non-isomorphic graphs). Hence learning {\em contextua}l based embeddings, {\em a la}  ELMo~\cite{peters2018deep}  namely, is crucial. Building upon the success of deep learning in images and words, graph neural networks (GNNs) have been recently developed for various graph learning tasks on graph-structured datasets. Most of existing GNNs are {\em task-specific} in the sense that   they are trained on datasets via supervised (or semi-supervised)  learning with task-specific labels, and thus the trained models cannot be directly applied to other tasks. In addition, they are often limited to learning {\em node embeddings}  	  (based on a node's local structure within an underlying graph), as opposed to {\em graph embedding} (defined as a single embedding for the whole graph), which is more general purpose, e.g.,  for graph classification or graph generation tasks. 
Besides graph embedding can \emph{serve as a node embedding} by operating on node's ego-graph.  

\begin{SCfigure}[\sidecaptionrelwidth][t!]
	
	\caption{\protect\rule{0ex}{4ex} Side figure shows two pairs of isomorphic graphs sampled from  real  but more interestingly different bioinformatics and quantum mechanic datasets namely {\fontfamily{cmr}\selectfont NCI1, MUTAG,  PTC, QM8}; suggesting the importance of learning universal graph embedding and performing transfer learning \& multi-tasking (for learning more generalized embeddings).}\label{fig:transfer}
	
	\begin{tikzpicture}[
	background rectangle/.style={fill=gray!20},
	]
	
	\centering
	\node[scale=0.75, inner sep=0pt] (fig1) at (1.4,1.4)  {\includegraphics[width=0.62\textwidth]{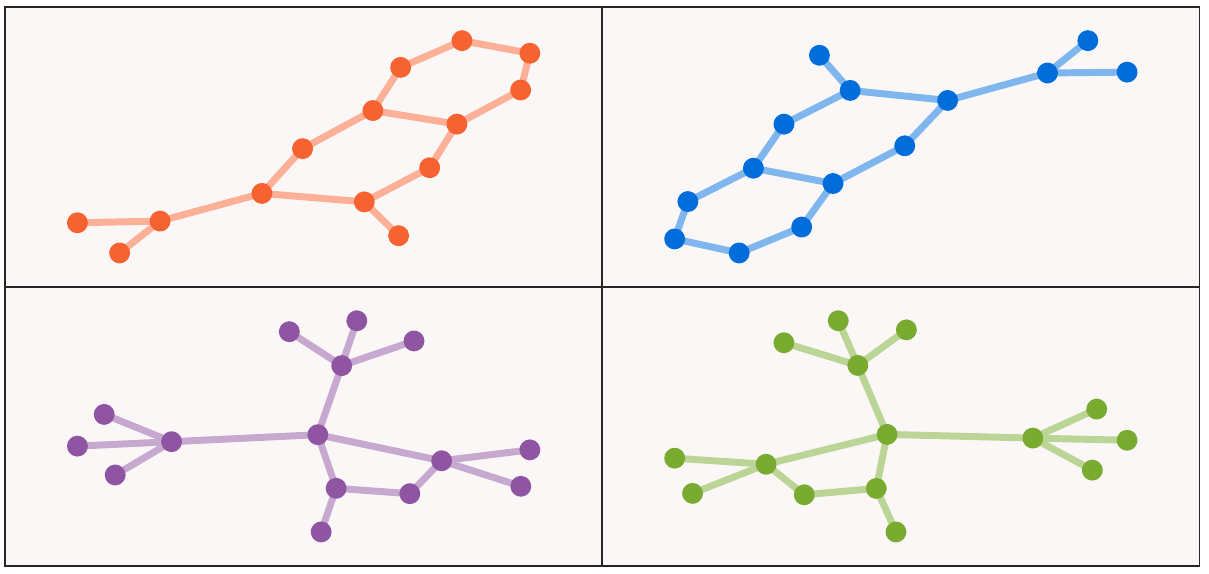}};
	
	\node[text width=1cm] at (-3.0, 4.2) {  {{\fontsize{9}{9}\selectfont \begin{center} MUTAG 	\end{center} }}};
	
	\node[text width=1cm] at (2.3, 3.9) { {{   \fontsize{9}{9}\selectfont    \begin{center}  NCI1 \end{center} }}};		
	
	\node[text width=1cm] at (-3.3, 1.0) {  {{\fontsize{9}{9}\selectfont \begin{center} QM8 	\end{center} }}};
	
	\node[text width=1cm] at (2.2, 1.0) { {{   \fontsize{9}{9}\selectfont    \begin{center}  PTC \end{center} }}};	
	
	\end{tikzpicture}
	
\end{SCfigure}
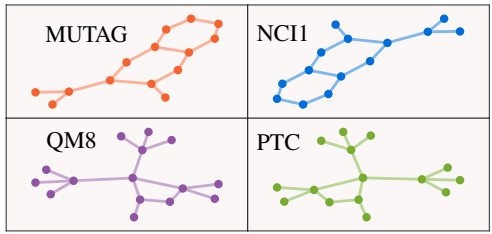

We envision a {\em deep universal graph  embedding neural network} (\textsc{DUGnn})	, which is  capable of the following: 1) It can be trained on diverse datasets (e.g., with different node features) for a variety of tasks in an {\em unsupervised} fashion to learn  {\em task-independent} graph embeddings; 2) The learned graph embedding model can be shared across different datasets, thus enjoying the benefits of   transfer learning; 3) The learned model can  further be adapted and improved for specific tasks using adaptive supervised learning.  Figure~\ref{fig:transfer} shows some sample graphs in four different real-world datasets from fields as diverse as bioinformatics and quantum mechanics. While the node features (and their meanings) can differ vastly but the underlying graphs governing them contain isomorphic graph structures. This   suggests that learning universal graph embeddings across diverse datasets is not only possible, but can potentially offer the benefits of transfer learning. From \emph{theoretical} point of view, we establishes the \emph{generalization guarantee} of  \textsc{DUGnn} model for graph classification task and discuss the \emph{role of transfer learning} in helping towards reducing the generalization gap. \emph{To best of our knowledge}, we are the first to propose doing \emph{transfer learning}    in the graph neural network domain.

In order to develop a universal graph embedding model, we need to overcome three main technical challenges.  Firstly, existing GNNs  operate at the {\em node feature-matrix} level.
Unlike images or words where the channels or embedding layer has a fixed input size,  in the context of graph learning, the initial node feature (a feature vector defined on nodes in a graph)   dimension can vary across different datasets. 
Secondly, the model complexity of GNNs is often  limited by the basic graph convolution operation, which in its purest form is the aggregation of neighboring node features and may suffer from the Laplacian smoothing problem~\cite{li2018deeper}. Lastly, the major technical hurdle is to devise an {\em unsupervised graph decoder} that is capable of regenerating or reconstructing the original graph directly from its graph embedding with minimal loss of information. 

We propose a \textsc{DUGnn} model   to tackle these challenges  with three carefully designed core components: 1) {\em Input Transformer}, 2) {\em Universal Graph Encoder}, and   3) {\em Multi-Task Graph Decoder}. 
Through extensive experiments and ablation studies, we show that the   \textsc{DUGnn}  model consistently outperforms both the existing state-of-art GNN models and Graph Kernels by an increased accuracy of $\mathbf{3\textbf{\%}-8\textbf{\%}}$ on  graph classification benchmark datasets.

\textbf{In summary, the major contributions of our paper are: }
\vspace{-0.5em}
\begin{itemize}[leftmargin=*]
	\setlength\itemsep{-0.1em}
	\item  We propose a novel theoretical guaranteed \textsc{DUGnn} model for universal graph embedding learning  that can be trained in {\em  unsupervised} fashion  and also capable of doing transfer learning. 
	\item  We leverage rich graphs kernels to design a multi-task graph decoder which incorporates the power of graph kernels in graph neural networks and get the best of both the worlds.  
	\item  Our \textsc{DUGnn}  model  achieves superior results in comparison with existing graph neural networks and graph kernels on  various types of graph classification benchmark datasets. 
\end{itemize}
\vspace{-1em}

\section{Related Work}~\label{sec:related_work}
\vspace{-2.5em}

Recently, various graph neural networks (viewed as   ``graph encoders'')   have been developed, in particular, learning task-specific node embeddings from graph-structured datasets. In contrast, we are interested in learning task-independent graph embedding. Furthermore, our problem also involves designing graph decoder -- graph reconstruction using    graph embedding -- which is far more challenging and  has not received much attention in the literature.  We overview the key related in these two areas along with graph kernels.  

\textbf{Graph Encoders and Decoders}: Under graph encoders, we consider both graph convolutional neural networks (GCNNs) and  message passing neural networks (MPNNs).  The early development of GNNs can be traced back to graph signal processing~\cite{shuman2013emerging} and cast in terms of  learning filter parameters  of the graph Fourier transform~\cite{bruna2013spectral, henaff2015deep}. Various GNN models have since been proposed~\cite{kipf2016semi, atwood2016diffusion, li2018adaptive, duvenaud2015convolutional,puy2017unifying, dernbach2018quantum, zhang2019quantum}  that   mainly attempt to improve the basic GNN model along two aspects: 1) enhancing the graph convolution operation  by developing novel  graph filters; 
and 2) designing   appropriate graph pooling operations. For instance, \cite{levie2017cayleynets, fey2018splinecnn}  employs complex graph filters  via Cayley and b-splines as a basis for the  filtering operation respectively. 
For graph pooling operations, pre-computed graph coarsening layers via the graclus multilevel clustering algorithm are employed in~\cite{defferrard2016convolutional}, while a differential pooling operation    is developed in \cite{ying2018hierarchical}. 
The authors in ~\cite{xu2018powerful} propose a sum aggregation pooling operation that is better justified in theory. 
The authors in~\cite{lei2017deriving, gilmer2017neural,dai2016discriminative, garcia2017learning} propose message passing neural networks (MPNNs), which are viewed as equivalent to GCNN models, as the underlying notion of  the graph convolution operation is the same. 
A similar line of studies have been developed in~\cite{hamilton2017inductive,velivckovic2018deep}   for learning node embedding on large graphs. 
In contrast, the authors in~\cite{zhang2018end, verma2018graph} propose GNNs for handling graph classification problem. 
The main limitations of all the aforementioned GNN models lie in that they must be trained from scratch on a new dataset and the embeddings are tuned based on a {\em supervised} (task-specific) objective function. As a result, learned embeddings are task dependent. There is  no mechanism for sharing the embedding model across different datasets. 
Unfortunately  work on designing graph decoder is currently under-explored    and partially falls under graph generation area~\cite{simonovsky2018graphvae, you2018graphrnn, de2018molgan, you2018graph}. But the graph generation  work   do not focus on recovering the exact graph structure but rather generating graph with similar characteristics. 


\textbf{Graph Kernels}: 
The literature on graph kernels is vast, we only outline a few. Some of the most popular graph kernels are Weisfeiler-Lehman   kernel~\cite{shervashidze2011weisfeiler}, graphlets~\cite{prvzulj2007biological,shervashidze2009efficient}, random walk or shortest path or anonymous walk based kernels~\cite{kashima2003marginalized,borgwardt2005shortest, ivanov2018anonymous}. Several graph kernels   based on more complex kernel functions have also been developed  that can capture sub-structural similarities at multiple levels; these include deep graph kernels~\cite{yanardag2015deep}, graph invariant  kernels~\cite{orsini2015graph} and  multiscale laplacian graph kernel~\cite{kondor2016multiscale}. Instead of directly computing graph kernels,  powerful  graph spectrum based methods have also been developed, for example,
Graphlet spectrum~\cite{kondor2009graphlet}   based on group theory,  and a family of graph spectral distances (FGSD)   based on spectral graph theory.

\vspace{-1em}
\section{Deep Universal Graph Embedding Model}\label{sec:model}
\vspace{-0.5em}
\noindent \textbf{Basic Setup and  Notations}: Let   $G=(V,E,\mathbf{A})$ be a graph   where $V$ is the vertex set, $E$ the edge set (with no self-loops) and $\mathbf{A}$ the       adjacency matrix, with  $N=|V|$ the graph size. We define the standard graph Laplacian $\mathbf{L}  \in \mathbb{R}^{N \times N} $ as $\mathbf{L}=\mathbf{D}-\mathbf{A} $, where $\mathbf{D}$ is the degree matrix. Let $\mathbf{X}\in \mathbb{R}^{N \times d}$ be the node feature matrix with $d$ as the input dimension and  $h$   denotes the hidden dimension.   Further let $f(\mathbf{L})$ be a function of the graph Laplacian i.e.,  $f(\mathbf{L})=\mathbf{U}f(\mathbf{\Sigma})\mathbf{U}^{T}$,  where $\mathbf{\Sigma}$ is the diagonal matrix  of  eigenvalues of $\mathbf{L}$ and $\mathbf{U}$ the eigenvector matrix. 

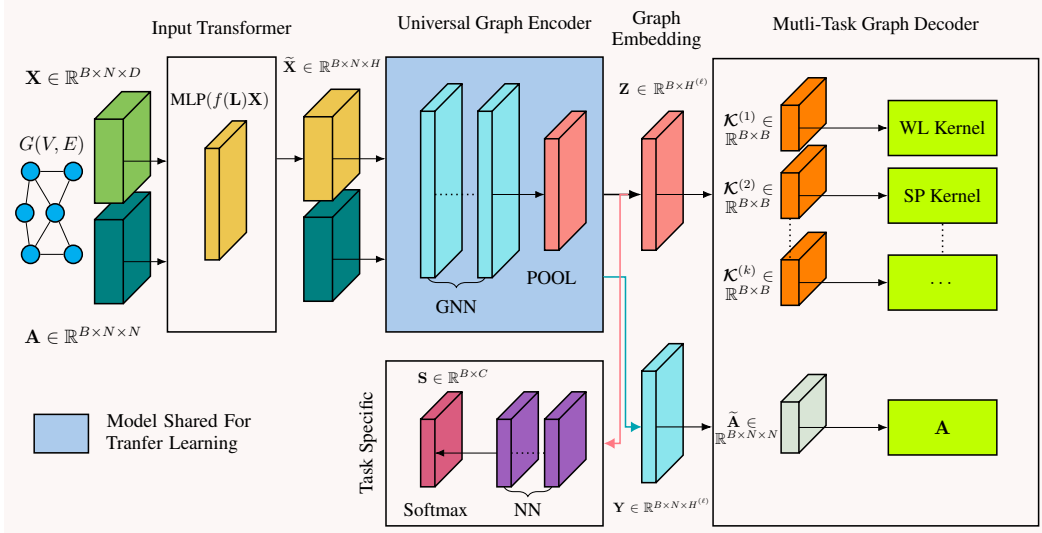
\begin{figure*}[t!]
	\centering
	\resizebox{14cm}{3.6cm}{
		
		\begin{tikzpicture}[
		background rectangle/.style={fill={rgb,255: red,252; green,247; blue,246}},
		show background rectangle
		]
		
		
		\node[parallelepiped, draw=black, thick,fill=teal , minimum width=0.5cm, minimum height=1.5cm, parallelepiped offset x=5mm]  (adjcube) at (0,0) {};
		\node[above  = 0.3cm of adjcube, parallelepiped, draw=black, thick, fill={rgb,255: red,135; green,196; blue,88}, minimum width=0.5cm,minimum height=1.5cm, parallelepiped offset x=5mm] (featcube) {};
		\node[above left  = 0.5 cm and -1cm of featcube, scale=1] (X) { $\mathbf{X} \in \mathbb{R}^{B\times N \times D}$}; 
		\node[below left = 0.3 cm and -1cm  of adjcube, scale=1] (A) { $\mathbf{A} \in \mathbb{R}^{B\times N \times N}$}; 
		
		
		\node[below left  = 0.05cm and 0.6cm of featcube, circle, draw=black, thick, fill=cyan, radius=1mm] (gnode1) {};
		\node[left  =   0.2cm of gnode1, circle, draw=black, thick, fill=cyan, radius=1mm] (gnode6) {};
		\node[above left  = 0.5cm and 0.2cm of gnode1, circle, draw=black, thick, fill=cyan, radius=1mm] (gnode2) {};
		\node[below left  = 0.5cm and 0.2cm of gnode1, circle, draw=black, thick, fill=cyan, radius=1mm] (gnode3) {};
		\node[above right  = 0.5cm and 0.08cm of gnode1, circle, draw=black, thick, fill=cyan, radius=1mm] (gnode4) {};
		\node[below right  = 0.5cm and 0.08cm of gnode1, circle, draw=black, thick, fill=cyan, radius=1mm] (gnode5) {};
		\draw[-] (gnode1) -- (gnode2);
		\draw[-] (gnode1) -- (gnode3);
		\draw[-] (gnode1) -- (gnode4);
		\draw[-] (gnode1) -- (gnode5);
		\draw[-] (gnode2) -- (gnode4);
		\draw[-] (gnode3) -- (gnode5);
		\draw[-] (gnode2) -- (gnode6);
		\draw[-] (gnode3) -- (gnode6);
		
		\node[above left = 0.8 cm and -0.8cm  of gnode1, scale=1] (nodetext) { $G(V, E)$}; 
		
		
		\node[below right = -4.5cm and 0.8cm of adjcube] (inputrect) [draw,  thick,minimum width=2cm,minimum height=5cm] {}; 
		\node[] (inputcube)  [parallelepiped, fill={rgb,255:  red,237; green,198; blue,80}, draw,thick, minimum width=0.1cm, minimum height=2cm] at ([xshift=-1em, yshift=-1em] inputrect.center) {}; 
		\node[above left = 0.6cm and -1.3cm of inputcube, scale=0.9] (inputext1) {$\text{MLP}(f(\mathbf{L})\mathbf{X})$}; 
		
		\draw[-Latex] (featcube.east)   (featcube) -- (inputrect.west |- featcube);
		\draw[-Latex] (adjcube.east)   (adjcube) -- (inputrect.west |- adjcube);
		\node[above  = 0.2cm of inputrect, scale=1] (H) {Input Transformer};
		
		
		
		\node[below right = -2.1cm and 0.5cm of inputrect, parallelepiped, draw=black, thick, fill=teal, minimum width=0.5cm, minimum height=1.5cm, parallelepiped offset x=5mm]  (adjcube2) {};
		
		\node[above  = 0.3cm of adjcube2, parallelepiped, draw=black, thick, fill={rgb,255:  red,237; green,198; blue,80}, minimum width=0.5cm,minimum height=1.5cm, parallelepiped offset x=5mm] (veccube) {};
		
		\node[above  right = 0.7cm and -1.3cm of veccube, scale=0.8, text width=3cm, align=center] (E) {$\mathbf{\widetilde{X}}  \in \mathbb{R}^{B\times N \times H}$ };
		
		\draw[-Latex] (inputrect.east) +(0mm,1.3)  |-    (veccube.west) ;
		
		
		
		\node[right = 2cm of inputrect] (encrect)  [draw=black, thick, minimum width=4cm,minimum height=5cm,fill={rgb,255: red,172; green,203; blue,236}] {}; 
		
		\draw[-Latex] (veccube.east)   (veccube) -- (encrect.west |- veccube);
		\draw[-Latex] (adjcube2.east)   (adjcube2) -- (encrect.west |- adjcube2);
		
		\node[] (gcnrect1)  [parallelepiped, fill={rgb,255:  red,139; green,229; blue,237}, draw,thick,minimum width=0.1cm,minimum height=3cm] at ([xshift=-7em]encrect.center) {}; 
		\node[] (gcnrect2)  [parallelepiped,  fill={rgb,255:  red,139; green,229; blue,237}, draw,thick,minimum width=0.1cm,minimum height=3cm] at ([xshift=-1em]encrect.center) {}; 
		\node[] (poolrect)  [parallelepiped,  fill={rgb,255:  red,250; green,139; blue,131}, draw,thick,minimum width=0.1cm,minimum height=2cm] at ([xshift=6em]encrect.center) {}; 
		\draw[thick, dotted] (gcnrect1) -- (gcnrect2);
		\draw[-Latex] (gcnrect2) -- (poolrect);

		\coordinate (a2) at (gcnrect1.south);
		\coordinate (a3) at (gcnrect2.south);
		\draw [decorate,decoration={brace,amplitude=5pt,mirror,raise=0ex}] (a2) -- (a3) node[midway,yshift=-1.5em]{GNN}; 
		\node[below = 0.25cm of poolrect] (pooltext1) {POOL}; 
		

		
		\node[right =   0.7cm  of encrect] (embrect) [ fill={rgb,255:  red,250; green,139; blue,131},draw=black, parallelepiped, thick,minimum width=0.1cm, minimum height=2cm] {}; 
		\node[above left = 0.7cm and -1.7cm of embrect, scale=0.8, text width=3cm, align=center] (E) {$\mathbf{Z} \in \mathbb{R}^{B\times H^{(\ell)}}$ };
		\node[above left = 1.5cm and -1.9cm of embrect, scale=1, text width=3cm, align=center] (Y) { Graph \\ Embedding }; 
		
		\node[below right =  0.7cm and 0.7cm  of encrect] (gcnnoutrect) [parallelepiped, draw=black, fill={rgb,255:  red,139; green,229; blue,237},
		thick,minimum width=0.1cm, minimum height=2cm] {}; 
		\coordinate (c1) at (gcnnoutrect.west);
		\draw[-{Latex[length=2mm, width=2mm]}, draw={rgb,255:  red,0; green,163; blue,177}, thick] (encrect.east) +(0mm, -30mm)  |- +(0.8, -3.0) |-  ($ (c1) + (0,-0.0) $) ;
		
		\draw[-Latex,  thick] (encrect) -- (embrect);
		
		\node[below left  = 0.2cm and -1.5cm  of gcnnoutrect, scale=0.7, text width=3cm, align=center] (E) {$\mathbf{Y} \in \mathbb{R}^{B\times N \times H^{(\ell)}}$ };
		
		
		\node[below right = -5cm and 2cm of encrect] (decrect)  [draw,thick,minimum width=6cm,minimum height=8.5cm] {};

		\draw[-Latex] (embrect.east)   (embrect) -- (decrect.west |- embrect);
		\draw[-Latex, line width=0.2mm] (gcnnoutrect.east)   (gcnnoutrect) -- (decrect.west |- gcnnoutrect);
		
		\node[above = 0.3cm of encrect] (enctext1) {Universal Graph  Encoder}; 
		\node[above = 0.3cm of decrect] (dectext1) {Mutli-Task Graph Decoder};

		\node[] (embkernelrect)  [parallelepiped, draw,thick, fill=orange, minimum width=0.3cm,minimum height=0.8cm] at ([xshift=-9em, yshift=17em]decrect.center) {};
		\node[] (embkernelrect2)  [parallelepiped, draw,thick, fill=orange, minimum width=0.3cm,minimum height=0.8cm] at ([xshift=-9em, yshift=10em]decrect.center) {};
		\node[] (embkernelrect3)  [parallelepiped, draw,thick, fill=orange, minimum width=0.3cm,minimum height=0.8cm] at ([xshift=-9em, yshift=1	em]decrect.center) {};
		
		\node[] (kernelrect1)  [draw,thick,minimum width=2cm,minimum height=1cm, fill=lime] at ([xshift=7em, yshift=17em]decrect.center) {WL Kernel}; 
		\node[] (kernelrect2)  [draw,thick,minimum width=2cm,minimum height=1cm, fill=lime] at ([xshift=7em, yshift=10em]decrect.center) {SP Kernel}; 
		\node[] (kernelrect3)  [draw,thick,minimum width=2cm,minimum height=1cm, fill=lime] at ([xshift=7em, yshift=1em]decrect.center) {$\dots$}; 
		
		
		
		\node[left = -0.05cm of embkernelrect, scale=0.9, text width=1.2cm, align=center] (E) {$\mathbfcal{K}^{(1)} \in \mathbb{R}^{B\times B}$ };
		\node[left = -0.05cm of embkernelrect2, scale=0.9, text width=1.2cm, align=center] (E) {$\mathbfcal{K}^{(2)} \in \mathbb{R}^{B\times B}$ };
		\node[left = -0.05cm of embkernelrect3, scale=0.9, text width=1.2cm, align=center] (E) {$\mathbfcal{K}^{(k)} \in \mathbb{R}^{B\times B}$ };
		
		\draw[-Latex] (embkernelrect.east) |- +(2, 0)  |-  (kernelrect1.west) ;
		\draw[-Latex] (embkernelrect2.east) |- +(2, 0)  |-  (kernelrect2.west) ;
		\draw[-Latex] (embkernelrect3.east) |- +(2, 0)  |-  (kernelrect3.west) ;
		
		\draw [dotted, thick] (kernelrect2.south) -- (kernelrect3.north);
		\draw [dotted, thick] (embkernelrect2.south) -- (embkernelrect3.north);
		
		
		\node[] (adjreconstrect)[parallelepiped,  fill={rgb,255:  red,217; green,229; blue,214}, draw,thick, minimum width=0.3cm,minimum height=0.9cm] at ([xshift=-9em, yshift=-14em]decrect.center) {};
		\node[] (Arect)  [draw, thick, minimum width=2cm, minimum height=1cm, fill=lime] at ([xshift=7em, yshift=-14em]decrect.center) {$\mathbf{A}$};
		\draw[-Latex] (adjreconstrect.east) --  (Arect.west) ;
		\node[  left =  0.13cm of adjreconstrect, scale=0.82, text width=1.2cm, align=center] (E) {$\widetilde{\mathbf{A}} \in \mathbb{R}^{B \times N\times N}$};
		
		

		
		\node[below = 0.5cm of encrect] (classrect)  [draw,thick,minimum width=4cm,minimum height=3cm] {}; 
		
		\coordinate (d1) at (classrect.east);
		\draw[-{Latex[length=2mm, width=2mm]}, draw={rgb,255:  red,255; green,120; blue,131}, thick] (encrect.east) +(9mm,0)  |- +(0.6, 0.0) |-  ($ (d1) + (0,0) $) ;
		
		\node[] (fcrect1)  [parallelepiped, draw,  fill={rgb,255:  red,158; green,95; blue,189}, thick,minimum width=0.1cm,minimum height=1.2cm] at ([xshift=6em, yshift=-1em]classrect.center) {}; 
		\node[] (fcrect2)  [parallelepiped, draw, fill={rgb,255:  red,158; green,95; blue,189}, thick,minimum width=0.1cm,minimum height=1.2cm] at ([xshift=1em, yshift=-1em]classrect.center) {};
		\node[] (outrect)  [parallelepiped, draw,  fill={rgb,255:  red,218; green,92; blue,127}, thick,minimum width=0.1cm,minimum height=1.2cm] at ([xshift=-7em, yshift=-1em]classrect.center) {}; 
		\coordinate (a4) at (fcrect1.south);
		\coordinate (a5) at (fcrect2.south);
		\draw [decorate,decoration={brace,amplitude=5pt,mirror,raise=0ex}]
		(a5) -- (a4) node[midway, yshift=-1.2em]{\text{NN}}; 
		\draw[dotted, thick] (fcrect1) -- (fcrect2);
		\draw[-Latex] (fcrect2) -- (outrect);
		\node[above left = 0.55cm and -1.5cm of outrect, scale=0.8, text width=2cm, align=center] (E) {$\mathbf{S} \in \mathbb{R}^{B \times C}$};
		\node[below left = 0.16cm and -1.0cm of outrect] (outtext1) {Softmax}; 
		\node[above left = -1.5cm and 0.6cm of classrect, rotate=90, anchor=north] (classtext1) {Task Specific};

		
		\node[below left = 2cm and 0.1cm of adjcube] (legend1) [draw, fill={rgb,255: red,172; green,203; blue,236}, thick, minimum width=1cm,minimum height=0.7cm] {}; 
		\node[right = 0.2cm of legend1, text width=3cm] (legend1text) {Model Shared For Tranfer Learning}; 
		
		

		\end{tikzpicture}
	}
	
	\caption{Above figure shows overall architecture of \textsc{Dugnn} model in the form of tensor transformations. Starting from the left, we have a graph $G(V, E)$ with node feature matrix $\mathbf{X}$ and adjacency matrix $\mathbf{A}$. $\mathbf{X}$ is first transform into a consistent feature dimension $\mathbf{\widetilde{X}}$ via Input Transformer. Next, Universal Graph Encoder computes graph embedding $\mathbf{z}$ and   output $\mathbf{Y}$  which is passed down to the decoder. Our Multi-task Graph Decoder  comprises of minimizing  graph kernel  losses and adjacency matrix reconstruction loss along with the optional supervised task  loss for joint end-to-end learning.
		\vspace{-1.5em}}
	\label{fig:architecture}
	
\end{figure*}

Figure~\ref{fig:architecture} depicts the overall architecture of  \textsc{DUGnn}  model.  We describe the core components namely 1) Input Transformer 2) Universal Graph Encoder 3) Multi-Task Graph Decoder, in more detail.

\vspace{-0.5em}
\subsection{Input Transformer}
\vspace{-0.5em}
The dimension of graph input features is task-specific and may differ across different datasets. To make our  universal graph encoder task-independent (or dataset-independent),  we   devise a   specific input layer which transforms  a given input node feature dimension $\mathbf{X}$ into a consistent  hidden feature dimension, i.e., $ \Ta: \mathbf{X} \in \mathbb{R}^{N \times d} \rightarrow \mathbf{\widetilde{X}} \in \mathbb{R}^{N \times h} $  and feed it to the universal graph encoder. 

It is well known that GNN  models  aggregate  feature information within a $K-$hop local neighborhood of a node~\cite{defferrard2016convolutional}. Hence these models heavily  rely on the  initial node features to capture crucial global structural information of a graph. However, in many datasets, the node features are absent (in other words, the inputs are merely the underlying graph structure). In these cases, many models choose $\mathbf{X}=\mathbf{I}$, the identity matrix.  Instead, we propose to initialize $\mathbf{X}$ as a  Gaussian random matrix (in absence of node features), which is justified by the following theorem.

\begin{theorem}[Graph Spectral Embedding Approx. with GNN Random   Feature Initialization]\label{theorem:gaussian_initialize} 
	Let $f(\mathbf{L})\in \mathbb{R}^{N \times N}$ be a function of graph Laplacian and $\mathbf{X}\in \mathbb{R}^{N \times d}  \sim \Na(0, \sigma^2)$ be a  Gaussian random matrix  initialize as a node feature (or embedding) matrix where $d \leq N$. 
	Then   $f(\mathbf{L})\mathbf{X}$  resultant embedding  is equal to a randomly projected  graph spectral embedding in $\mathbb{R}^{d}$ space i.e, $f(\mathbf{L})\mathbf{X} = \Ra(\mathbf{U}f(\mathbf{\Sigma}))$ where $\Ra(\cdot)$ is some random projection in $\mathbb{R}^{d}$ space.
\end{theorem}

\noindent\textbf{{Remarks}}: The proof relies on the fact that the projection of a Guassian random matrix $\mathbf{X}$  onto an eigenspace $\mathbf{U}$ of $\mathbf{L}$  preserves  the Gaussian properties of $\mathbf{X}$~\cite{paratte2016fast} with probability 1. One of the consequences of Theorem~\ref{theorem:gaussian_initialize} is that  we can approximate different spectral dimension reduction techniques with suitably chosen $f(\cdot)$ function. For instance,  if we choose $f(\cdot)$ as the identity function, the transformation becomes the  Laplacian eigenmaps~\cite{belkin2003laplacian}. 
Moreover,  remarkably $f(\mathbf{L})\mathbf{X}$ is nothing but a graph convolution operation and thus provides an approximation of the graph spectral embedding. This provides a theoretical explanation for the competitive performance of a randomly initialized GNN model as previously noted in~\cite{kipf2016semi, velivckovic2018deep}. 

With  	   appropriately initialized  input feature vector $\mathbf{X}$,  our input  transformer performs the following  operation, $\Ta(\mathbf{X})  = \text{MLP}\Big( f(\mathbf{L})\mathbf{X}\Big)$
where MLP is a multi-layer perceptron  neural network.
\vspace{-0.8em}
\subsection{Universal Graph  Encoder}
\vspace{-0.7em}
Our universal graph encoder is based on a GNN model but with several key improvements. Let $g(\mathbf{L})$ be a graph filter function. Some early adopted graph filters are  polynomial functions of $\mathbf{L}$~\cite{defferrard2016convolutional}. We simply choose $g(\mathbf{L}) = \mathbf{D}^{-1/2}\mathbf{A}\mathbf{D}^{-1/2} + \mathbf{I}$ as  in~\cite{kipf2016semi}. 
The  $\ell^{th}$ layer output of a GNN model  can be written in general as, $\Fa^{(\ell)}(\mathbf{\widetilde{X}}, \mathbf{L})   = \text{MLP}  \Big( g(\mathbf{L})\Fa^{(\ell-1)}(\mathbf{\widetilde{X}}, \mathbf{L}) \Big)$ where $\Fa^{(0)}(\mathbf{\widetilde{X}}, \mathbf{L}) =\mathbf{\widetilde{X}}$ and $\Fa^{(\ell)}(\mathbf{\widetilde{X}}, \mathbf{L}) \in \mathbb{R}^{N \times h^{(\ell)}}$. 
To further improve the model complexity of our graph encoder, we  capture  the  higher order  statistical  moment information of features during the graph convolutional operation, similar to graph capsule networks~\cite{verma2018graph} as follows,   



\vspace{-2em}
\begin{equation}\label{eq:capsule}
\begin{split}
\Fa^{(\ell)}(\mathbf{\widetilde{X}}, \mathbf{L})  &  = \text{MLP}\Bigg(   \sum_{p=1}^{P} \text{MLP}\Big(  g(\mathbf{L})\big(\Fa^{(\ell-1)}(\mathbf{\widetilde{X}}, \mathbf{L}) \big)^{p}\Big)  \Bigg)\\
\end{split}
\end{equation}
\vspace{-2em}

where   $P$ is the number of instantiation parameters. 
Other than boosting the model complexity,  there is standing problem of Laplacian smoothing~\cite{li2018deeper} in GCNN models. To overcome the smoothing problem, we propose to concatenate the output of intermediate   encoding layers and feed it into subsequent layers. This way our graph encoder can learn on multi-scale smooth features of a node and thus can avoid under-smoothing or over-smoothing issues in a GCNN based model. Fortunately, it also enjoy the side benefits of   alleviating   vanishing-gradient problem and strengthen feature propagation in deep networks as shown in  Dense-CNN models~\cite{huang2017densely}. To obtain  the final graph embedding $\mathbf{z}\in \mathbb{R}^{h^{(\ell)}}$, we perform   sum-pooling on all nodes $\mathbf{z} = \sum_{i=0}^{N} \Fa_{i}^{(\ell)}(\mathbf{\widetilde{X}}, \mathbf{L}) \in \mathbb{R}^{h^{(\ell)}}$ and show its representational power below. 

\begin{theorem}[Deep Universal Graph Embedding Representational Power]\label{theorem:rep_power} 
	Deep Universal Graph Embedding model initialized with  node features as  graph spectral embeddings  is   \textbf{atleast} as powerful as the classical Weisfeiler-Lehman (WL) graph isomorphism test. \end{theorem}
\vspace{-0.5em}
\noindent\textbf{{Remarks}}: Theorem~\ref{theorem:rep_power}'s subpart where \textsc{DUGnn} representation power is \emph{same} as  WL graph isomorphism test  directly follows from  Theorem 3 of the paper~\cite{xu2018powerful}. To make \textsc{DUGnn} more powerful  than the classical WL graph isomorphism test (where nodes features are either  identical or equal to respective node degree), one   can initialize \textsc{DUGnn} node features with graph spectral embeddings. As a result,   \textsc{DUGnn} can now differentiate certain regular graphs where the classical  WL graph isomorphism test fails.  The main trick here is to initialize node features such they  depend on the full graph structure rather than the local structure.  Coincidently, we have Theorem~\ref{theorem:gaussian_initialize} to approximate graph spectral embedding in our \textsc{DUGnn} model and that too in fast manner. Although in all fairness, the same trick can also make   WL more powerful. 


Next, we provide the generalization guarantee of  \textsc{DUGnn} model and further discuss  the role of transfer learning in   reducing the generalization gap.
\begin{theorem}[Deep Universal Graph Embedding Model Generalization Guarantee]\label{thm:gcnn_gen_bound} \textit{Let $A_S$ be a  single layer Universal Graph Encoder        equipped with the graph convolution filter $g(L)$  and trained on a dataset  $S$  using  the SGD algorithm for $T$ iterations. Let the loss \& activation functions be Lipschitz-continuous and smooth. Then the  following expected generalization guarantee   holds with probability at least  $1 -\delta $ having $\delta \in (0,1)  $,}
	\vspace{-0.5em} 	
	\begin{equation*} 	
	\begin{split}
	\mathbf{E}_{\textsc{sgd}}[R(A_S) ] &  \leq  \mathbf{E}_{\textsc{sgd}}[R_{emp}(A_S)] +  \BigO \Big( P  N^{T+1}(\lambda_{G}^{\max})^{2T}\Big) \Bigg(\frac{1}{m} + \sqrt{\frac{\log \frac{1}{\delta}}{2m}}\Bigg) + C\sqrt{\frac{\log \frac{1}{\delta}}{2m}} \\
	\end{split}	 
	\end{equation*} 
	\vspace{-0.5em}
	
	where   $\mathbf{E}_{\textsc{sgd}}[R(\cdot)]$ is the expected risk taken over   the randomness due to SGD,   $R_{emp}(\cdot)$ is the empirical risk, $m$ is the number of training graph samples, $N$ is the maximum graph-size, $\lambda_{G}^{\max}$ is the largest eigenvalue of graph filter  and $C$  is a upper bound on the loss function.
\end{theorem} 
\vspace{-0.5em}
\noindent\textbf{{Remarks}}: Theorem~\ref{thm:gcnn_gen_bound} relies on showing the fact that Universal Graph Encoders are uniformly stable~\cite{bousquet2002stability} and  has several implications. First, normalized graph filters   are   theoretically more stable (besides numerically) since  $\lambda_{G}^{\max} \leq 1$ and parameter $P$ controls the classic bias-variance tradeoff.  Also the theorem   does not bear any restrictions on type of graph datasets employed for training and establishes that transfer learning remains beneficial between different datasets for graph classification task. Intuitively GNN parameters are learned based  on   the local ($k-$hop) graph-structure and having more samples will always help   towards generalizing better on unseen data and    reduces the generalization error at a rate $\BigO(\frac{1}{\sqrt{m}})$.

\vspace{-0.8em}
\subsection{Multi-Task Graph Decoder}
\vspace{-0.7em}
Multi-task learning     have shown to yield superior results in natural language processing~\cite{mccann2018natural, devlin2018bert}. We want to equip our  \textsc{DUGnn}  model with built-in capabilities of multi-task learning. For this, we employ multiple prediction metrics in our graph decoder to enable it to learn more generalized graph embedding useful for different learning tasks. Note that supervised task specific  loss i.e.,  cross-entropy loss $\La_{\text{class}}$  is not considered as the part of our multi-task decoder (see Figure~\ref{fig:architecture}).

\noindent \textbf{Graph Adjacency Matrix Reconstruction Loss}: The major technical challenge in devising a general purpose graph decoder is reconstructing 
the original graph structure {\em directly from its graph embedding} vector $\mathbf{z}$. 
We consider minimizing adjacency reconstruction loss  $\La_{A}$ as the first task of our graph decoder. Following adjacency loss  $\La_{A}$ is incurred during mini-batch process, $\La_{A} =   \lambda_A\sum_{i=1}^{B} \ell_{\textsc{CE}}\big(\sigma(\mathbf{Y}_i \mathbf{Y}_i^{T}), \mathbf{A}_{i}\big)$
where $\mathbf{Y} = \Fa^{(\ell)}(\mathbf{\widetilde{X}}, \mathbf{L})$ is the encoder output, $\ell_{\text{CE}}$ is binary cross entropy loss corresponding to presence or absence of each edge and $\lambda_A$ is loss weight.  There are two shortcomings of this task. First, $\La_A$   does not take  graph embedding $\mathbf{z}$ directly into account and as such   graph embedding may incur significant loss of information after the pooling operation. Second,  with $O(N^2)$ computing $\La_{A}$ (in every batch iteration) may become   expensive on datasets with large graph-size. In our experiments,  we were able to compute adjacency  reconstruction loss on all  datasets except D\&D.



\noindent \textbf{Graph Kernel Loss}: We propose a novel solution which leverages rich graph kernels to overcome the shortcomings present in  the  first task. In essence, graph kernels are developed to capture various key sub-structural properties of a graph: for two graphs, $\Ka(G_i,G_j)$ provides a measure of their similarity. 
In our decoder design, we incorporate multiple graph kernels \{$\Ka^{(k)}$\}  
to jointly learn and predict the quality of (universal) graph 
embedding $\mathbf{z}$ directly by minimizing the following unsupervised  joint graph kernel loss  $\La_{\Ka}^{(\text{unsup})}$, 

\vspace{-1.5em}
\begin{equation*}\label{eq:gcnn_op}
\begin{split}
\ell_{\Ka^{(k)}} =&  \sum_{i=1}^{B}\sum_{j=1}^{B}\ell_{\textsc{MSE}}\Big(\sigma\big(\mathbf{z}_i^{T} \mathbf{W}_{k}\mathbf{z}_j \big) , \Ka_{ij}^{(k)} \Big), \hspace{2em}
\La_{\Ka}^{(\text{unsup})} =      \lambda_{\Ka}\sum_{k=1}^{K} \lambda_k \ell_{\Ka^{(k)}} \\ 
\end{split}
\end{equation*}
\vspace{-1.5em}

where $\ell_{\text{MSE}}$ is the mean square error loss, $\sigma\big(\mathbf{z}_i^{T} \mathbf{W}_{k}\mathbf{z}_j \big)$ is a (learned) similarity function between two graph embeddings and  $\mathbf{W}_{k} \in \mathbb{R}^{h^{(\ell)} \times h^{(\ell)}}$ is the associated   kernel weight parameter.  By leveraging precomputed graph kernels, our computation is cheap (in every batch iteration), but also  take the graph embedding  $\mathbf{z}$ directly into account for the joint learning of our graph encoder-decoder model.  


\noindent \textbf{Adaptive Supervised Graph Kernel Loss}:  For supervised task learning problems, we augment the   loss function   with an adaptive supervised graph kernel loss function. Specifically, we focus on graph kernels that are aligned with the task objective. As shown in Equation (\ref{eq:loss_adpative_kernel}), if class labels of two graphs are the same, then we choose the graph kernel with maximum similarity value and vice-versa. Thus, we are making sure to pick only those sub-structural properties that are relevant to a given specific task. For instance, in MUTAG dataset,  counting the number of cycles is more important than computing the  distribution of random walks (see appendix for more details).

\vspace{-1.5em}
\begin{equation}\label{eq:loss_adpative_kernel}
\begin{split}
\La_{\Ka}^{(\text{sup})}  & =  \lambda_{\Ka}\sum_{i=1}^{B}\sum_{j=1}^{B}   
\ell_{\text{MSE}}\Big(\sigma\big(\mathbf{z}_i^{T} \mathbf{W}_{k}\mathbf{z}_j \big),    
I_{(y_i =y_j)}\max_{k}(\Ka_{ij}^{(k)}) +   \big(1-I_{(y_i =y_j)} \big) \min_{k}(\Ka_{ij}^{(k)}) \Big) 
\end{split}
\raisetag{4\normalbaselineskip}
\end{equation}


\noindent \textbf{Computational Complexity}: Precomputed graph kernel representation are typically bounded by $O(N^2)$ time per graph (examples include WL, FGSD kernel) where $N$ is number of nodes in a graph. Further computing graph kernel in each batch iteration requires  $O(B^2)$ time and space  where $B$ is number of graphs in batch size. As a a result, our model time and space complexity is bounded by $O(B^2N^2)$ in each batch iteration.

\section{Experiment and Results}\label{sec:exp_results}

We evaluate  \textsc{DUGnn}   thoroughly on a variety of graph classification benchmark datasets.

\renewcommand{\arraystretch}{1.8}
\begin{table*}[t!]
	\centering
	\fontsize{5.5}{6.5}\selectfont

	\resizebox{\textwidth}{!}{	\begin{tabular}{   >{\raggedright}p{1cm}  !{\vrule width0.8pt}      K{1cm}  !{\vrule width0.8pt} K{1cm}  !{\vrule width0.8pt} K{1cm} !{\vrule width0.8pt} K{1cm}   !{\vrule width0.8pt} K{1cm}   !{\vrule width0.8pt}K{1cm}   !{\vrule width0.8pt} K{1cm} !{\vrule width0.8pt} K{1cm}  !{\vrule width0.8pt} K{1cm} | }
			
			\multirow{1}{*}{\hspace{-1em}\textbf{Model / Dataset}} &      \multicolumn{1}{c!{\vrule width0.8pt}}{{  PTC}} &	\multicolumn{1}{c!{\vrule width0.8pt}}{ {  PROTEINS}} &	 \multicolumn{1}{c!{\vrule width0.8pt}}{ {  ENZYMES}} &  \multicolumn{1}{c!{\vrule width0.8pt}}{ {  D\&D}*} &  \multicolumn{1}{c!{\vrule width0.8pt}}{ {  NCI1}} &	 \multicolumn{1}{c!{\vrule width0.8pt}}{  {  COLLAB}}  &	 \multicolumn{1}{c!{\vrule width0.8pt}}{  {  IMDB-B}} &	 \multicolumn{1}{c!{\vrule width0.8pt}}{  {  IMDB-M}}  \\ 

			\hspace{-1em}{\textbf{(\#Graphs)}} 						&     {$344$} 					&	 {$1113$} 				&	 {$600$} 					&	 {$1178$} 					&  {$4110$} 					& 	{$5000$}  &  {$1000$} &	 {$1500$} \\  
			\hspace{-1em}\multirow{1}{*}{\textbf{(Avg. \#nodes)}}    &  {$25.56$} 				 & {$39.06$}   					&	 {$32.60$} 					&	 {$284.32$} 				&  {$29.80$} 					& 	 {$ 74.49 $}  &  {$ 19.77 $} &	 {$ 13.00 $}	\\  	
			\hspace{-1em}\multirow{1}{*}{\textbf{(Max. \#nodes)}} 	&    {$109$} 				 &	 {$620$} 					&	 {$126$}  					&	 {$5748$}					&  {$111$} 						& 	{$492$}  &  {$136$} &	 {$89$}	 \\  \Xhline{2\arrayrulewidth}
			\hspace{-1em}{RW}[\citeyear{gartner2003graph}]       			&$57.8 \pm 1.3$            &  $74.2 \pm 0.4$  			&  $24.1\pm 1.6$ 			& $>24$ hrs 					&$>24$ hrs						&  $>24$ 						& $>24$ & $>24$ \\  \hline
			\rowcolor[RGB]{252, 247, 246}  
			\hspace{-1em}{SP}[\citeyear{borgwardt2005shortest}]       		&$58.2 \pm 2.4$            &  $75.0 \pm 0.5$ 			&  $40.1 \pm 1.5$   			&$>24$hrs						&$73.0\pm0.2$					&  --- 			  	&---&---  \\  \hline
			
			\hspace{-1em}{GK}[\citeyear{shervashidze2009efficient}]       	&$57.2 \pm 1.4  $          &  $71.6 \pm 0.5$  			&  $26.6 \pm 0.9$  			&$78.4 \pm 1.1$ 				&$62.2 \pm 0.2 $				&  $72.8 \pm 0.2 $ 			&$65.8 \pm 0.9$&  $43.8 \pm 0.3 $ 	 \\  \hline
			\rowcolor[RGB]{252, 247, 246}  
			\hspace{-1em}{WL} [\citeyear{shervashidze2011weisfeiler}]   	&$57.9 \pm 0.4$            &  $74.6 \pm 0.4$  			&	  $52.2 \pm 1.2$  		&$ {79.7 \pm 0.3}$  			&$82.1 \pm 0.1$				&   $78.9 \pm 1.9$
			& $73.8 \pm 3.9$	&  $50.9 \pm 3.8 $  \\  \hline
			
			\hspace{-1em}{DGK}[\citeyear{yanardag2015deep}]   				&$60.0 \pm 2.5 $           &  $75.6 \pm 0.5$ 			&  $53.4 \pm 0.9$ 			& $73.5\pm 1.0$  			&$80.3 \pm 0.4$				&   $73.0 \pm 0.2$  & $66.9\pm0.5$ &  $44.5\pm0.5$   \\  \hline
			\rowcolor[RGB]{252, 247, 246}  
			\hspace{-1em}{MLG}[\citeyear{kondor2016multiscale}]        		&$63.2 \pm 1.4  $          &  $76.3 \pm 0.7 $  			&  $61.8 \pm 0.9  $   		&$78.1 \pm 2.5 $				&$81.7 \pm 0.2$				&  --- 			&---&	---   \\  \hline				
			\hspace{-1em}{FSGD}[\citeyear{verma2017hunt}]        			& $62.8$ 					 &  $73.4$ 					&  --   						& $77.1$						& $79.8$ 						&  $80.02$  & $ 73.62$ &  $52.41$ 	  \\  \hline
			\rowcolor[RGB]{252, 247, 246}  
			\hspace{-1em}{AWE}[\citeyear{ivanov2018anonymous}]              & --- 						 &   ---  						&  $35.7 \pm 5.9$ 			& $71.5 \pm 4.0$				& --- 							&   $73.9\pm 1.9$  & $74.4 \pm 5.8$&  $51.5 \pm 3.6$    \\   \Xhline{2\arrayrulewidth}

			
			
			\hspace{-1em}{PSCN}[\citeyear{niepert2016learning}]      		&$62.2 \pm 5.6$		     &  $75.0 \pm 2.5$ 			&   ---  						&    --- 						&$76.3 \pm 1.6$ 				&   $72.6\pm 2.1$  &$71.0 \pm 2.2 $&  $45.2\pm2.8$ 	 \\  \hline
			\rowcolor[RGB]{252, 247, 246}  
			\hspace{-1em}{DCNN}[\citeyear{atwood2016diffusion}]      		&$56.6 \pm 2.8$			 &  $61.2 \pm 1.6 $  			&  $42.4\pm 1.7$  			&$58.0\pm 0.5$				&$56.6 \pm 1.0$				&  $52.1 \pm 0.7$  &$49.0 \pm 1.3 $&  $33.4 \pm 1.4$  \\  \hline
			\hspace{-1em}{ECC}[\citeyear{simonovsky2017dynamic}]     		& --- 						 &  --- 						&  $45.6$  					&$72.5$						&$76.8$						&  --- 					&---&---	  \\  \hline
			\rowcolor[RGB]{252, 247, 246}  
			\hspace{-1em}{DIFF}[\citeyear{ying2018hierarchical}] 		& --- 					     &  $ 76.2 $  					&  $ 62.5 $    					& $ 80.6$ 						& --- 							&   $82.13 $						& ---&---	 \\  \hline
			\hspace{-1em}{DGCN}[\citeyear{zhang2018end}]       	  		&$58.5 \pm 2.4$			 &  $75.5 \pm 0.9 $  			&   $51.0 \pm 7.2$  		  	&${79.3\pm0.9}$ 				&$74.4\pm0.4$					&       $73.7 \pm 0.4 $  &$70.0\pm 0.8$&  $47.8\pm0.8$ 	\\  \hline
			\rowcolor[RGB]{252, 247, 246}  
			\hspace{-1em}{GIN}[\citeyear{xu2018powerful}]       	  		&$64.6 \pm 7.0$            &  $76.2 \pm 2.8 $ 			&   ---    						& ---							& $82.7 \pm 1.7$  			&   $ 80.2 \pm 1.9 $&  $75.1 \pm 5.1 $   & $52.3 \pm 2.8$  	 \\  \hline		
			\hspace{-1em}{GCAPS}[\citeyear{verma2018graph}]          		&$ 66.0\pm 5.9 $           &  $ 76.4\pm 4.1 $ 			&  $ 61.8 \pm 5.3 $   	   	& $ 77.6 \pm 4.9 $ 			&$  82.7 \pm 2.3 $			& $77.7 \pm 2.5$  &$71.6 \pm 3.4$&  $48.5 \pm 4.1$  	  \\   \hline	
			\rowcolor[RGB]{252, 247, 246}  
			\hspace{-1em}{\scriptsize{\textbf{\textsc{DUGnn}}}}   		  		&${\mathbf{74.7\pm 6.0}}$  &  $ \mathbf{81.7 \pm 2.4}$  & $ \mathbf{67.3 \pm 4.8}$   	&  $ \mathbf{82.4 \pm 3.3}$   & $ \mathbf{85.5 \pm 1.2}$   	& $\mathbf{84.2 \pm  2.7}$    	& $\mathbf{78.7 \pm 4.9}$ & $\mathbf{56.1 \pm 2.3}$   \\  \hline
			
	\end{tabular}}
	
	\caption{Graph classification  accuracy   on bioinformatics and social network datasets. 
		Result in \textbf{bold} indicates the  best reported accuracy. Top half of the table compares results  with  Graph Kernels    while bottom half compares results with Graph Neural Networks.  *On D\&D dataset, we omit computing adjacency reconstruction loss due to GPU memory constraints.} 
	\label{table:bio_results}
	\vspace{-1em}
\end{table*}

\noindent \textbf{\textsc{DUGnn} Model Configuration}: In the Input Transformer,  we set the hidden dimension  to $h\in\{16, 32, 64\}$. For  the graph datasets without node features, we initialize $\mathbf{X}$ using a Gaussian random matrix and  choose $f(\mathbf{L})$ as the normalized symmetric   Laplacian.
For the Universal Graph Encoder, we build $\ell\in\{5, 7\}$ layer deep GNN network with  the internal hidden dimensions chosen from $h^{(\ell)}\in\{16, 32, 64\}$ and pick the  instantiation parameter from $p\in \{1, 2, 4\}$.   In the  Multi-Task Decoder,  we employed  three graph kernels loss besides adjacency reconstruction loss  1) WL-Subtree   2) Shortest-Path     3) FGSD,   as they are relatively fast to compute and all decoder losses are added with equal weights.   To keep the network outputs stable, we use   batch normalization between   layers   along with the  $L2$ weight norm regularization and dropout  techniques to prevent overfitting.
In addition, we employed the ADAM optimizer with varying learning rates as proposed in~\cite{vaswani2017attention} and its parameters are set as: the max epoch  $3000$, warmup epoch $2$, initial learning rate $10^{-4}$, max learning rate  $10^{-3}$ and final learning rate $10^{-4}$. All models are trained with early stopping criteria based on validation loss.


\noindent \textbf{Datasets and Baselines}:  We employed  $5$  bioinformatics and $3$ social network benchmark  datasets to evaluate the \textsc{DUGnn} model on the graph classification  task, namely,   {\fontfamily{cmr}\selectfont 
	PTC, PROTEINS, NCI1, D\&D, ENZYMES, COLLAB, IMDB-BINARY} and {\fontfamily{cmr}\selectfont IMDB-MULTI} and further details are present in~\cite{yanardag2015deep}. 
We compare the \textsc{DUGnn} model performance against $7$ recently proposed   GNNs and $8$   state-of-art Graph Kernels as shown in Table~\ref{table:bio_results}. 

\noindent \textbf{Experimental Set-up}: We first train \textsc{DUGnn} in an unsupervised fashion on   all mentioned datasets together. Here the universal graph encoder   is shared across all the datasets. Next, we fine tune our  \textsc{DUGnn} for each dataset separately using a  cross-entropy loss ($\La_{\text{class}}$) corresponding to the graph labels and the adaptive supervised kernel loss.  We report  $10$-fold cross validation results obtained by closely following the same experimental setup used in previous studies~\cite{ying2018hierarchical, xu2018powerful}.
To make a fair comparison, we either cite  the best cross validation results  previously reported   or run the source code if available according to author's guidelines.  
Further   details  are   present in the supplementary. 

\noindent \textbf{Graph Classification Results}: Table~\ref{table:bio_results} shows the classification   results on  considered   datasets based on graph neural network models and graph kernel methods. It is clear that \textsc{DUGnn}  \textbf{consistently outperforms} every state-of-art  GNN model  by a margin of  $\mathbf{3\textbf{\%}-8\textbf{\%}}$ increase in prediction accuracy on \emph{both bioinformatics  as well as  social-network datasets} (with the highest accuracy achieved  on the {\fontfamily{cmr}\selectfont PTC}).
On relatively  small datasets such as {\fontfamily{cmr}\selectfont PTC \& ENZYMES},   the increase  in accuracy is around $\mathbf{6\textbf{\%}-8\textbf{\%}}$. This empirically confirms our hypothesis that transfer learning in the graph domain  is quite   beneficial,  especially where available graph samples are limited. 

Our \textsc{DUGnn} model  also  significantly outperforms all the state-of-art graph kernel methods. We again  observe a consistent performance gain of   $\mathbf{3\textbf{\%}-11\textbf{\%}}$  (again with the highest increase on the  {\fontfamily{cmr}\selectfont PTC} dataset). Interestingly, our \textsc{DUGnn} integrated with graph kernels in the multi-task decoder \emph{outperforms the  WL-subtree  kernel, FGSD and SP  kernel}.

\section{Ablation Studies and Discussion}\label{sec:ablation_study}

We now take a closer look at the performance of each component of \textsc{Dugnn} model by   performing various ablation studies and show their individual importance and contributions. 

\noindent \textbf{How powerful is our    Universal Graph Encoder without multi-tasking and transfer learning?}

\renewcommand{\arraystretch}{2}
\begin{SCtable}[\sidecaptionrelwidth][h!]
	\centering
	\fontsize{7}{8}\selectfont
	\begin{tabular}{  p{2.5cm} |     K{2cm}  !{\vrule width0.8pt} K{2cm} !{\vrule width0.8pt} K{2cm}   | }
		
		\hspace{-0.8em}	\multirow{1}{*}{\textbf{Model / QM8 Results}} &       	\multicolumn{1}{c!{\vrule width0.8pt}}{Val MAE ($\times 10^{-3}$)}  &  \multicolumn{1}{c!{\vrule width0.8pt}}{Test MAE ($\times 10^{-3}$)}      \\ \hline
		
		
		
		\hspace{-0.8em}{\fontfamily{cmr}\selectfont MPNN}[~\citeyear{gilmer2017neural}]   		   &     $14.60$ 	    &  $14.30$ \\  \hline
		\rowcolor[RGB]{252, 247, 246}  
		\hspace{-0.8em}{\fontfamily{cmr}\selectfont DTNN}[~\citeyear{schutt2017quantum}]  		   &     $17.00$ 	    &  $16.90$ \\  \hline
		\hspace{-0.8em}{\fontfamily{cmr}\selectfont GCNN}[~\citeyear{wu2018moleculenet}]       	   &     $15.00$ 	    &  $14.80$ \\  \hline
		\rowcolor[RGB]{252, 247, 246}  
		\hspace{-0.8em}{\small{\textbf{\textsc{DUGnn}}}}   	- $\La_{A}$ - $\La_{\Ka}$ &     $\mathbf{11.16}$  	    &  $\mathbf{11.54}$ \\  \hline

		\Xhline{2\arrayrulewidth}
	\end{tabular}	
	\caption{\protect\rule{0ex}{4ex} Ablation Study of   Universal Graph Encoder on quantum mechanics dataset.  \textsc{DUGgnn} - $\ell_{A}$ - $\ell_{\Ka}$   (model trained from scratch without multi-task decoder) sets the new state-of-art result on {\fontfamily{cmr}\selectfont QM8} dataset. } 
	\label{table:ugd_perform}
	\vspace{-3em}
\end{SCtable}

We conduct an experiment where the \textsc{Dugnn} model is trained from \emph{scratch without the multi-task decoder}  and evaluate its performance on a   large quantum mechanic  dataset {\fontfamily{cmr}\selectfont QM8} (containing $21786$ compounds). For this purpose, we use the same experimental settings provided in~\cite{wu2018moleculenet} including the same  datasplits  given by the {\fontfamily{cmr}\selectfont DeepChem}\footnote{https://deepchem.io/} implementation, and compare it against the three state-of-the-art GNNs: 1) MPNN~\cite{gilmer2017neural}, 2) DTNN~\cite{schutt2017quantum} and 3) GCNN\footnote{\label{note1} http://moleculenet.ai/models}. From Table~\ref{table:ugd_perform}, it is clear that our universal graph encoder significantly outperforms these GNNs by a large margin of  $\mathbf{20\textbf{\%}-30\textbf{\%}}$  in terms of the mean absolute error (MAE), achieving the new state-of-art result on the {\fontfamily{cmr}\selectfont QM8} dataset.

\noindent \textbf{How much gain do we see from sharing pretrained \textsc{Dugnn} model via transfer learning?}

\renewcommand{\arraystretch}{2}
\begin{SCtable}[\sidecaptionrelwidth][h!]
	\centering
	\fontsize{7}{8}\selectfont
	\begin{tabular}{  p{3.0cm} |     K{1.6cm}  !{\vrule width0.8pt} K{1.6cm} !{\vrule width0.8pt} K{1.6cm}   | }
		
		\hspace{-0.8em}	\multirow{1}{*}{\textbf{Model / Dataset}} &       	\multicolumn{1}{c!{\vrule width0.8pt}}{{\fontfamily{cmr}\selectfont NCI1}}  &  \multicolumn{1}{c!{\vrule width0.8pt}}{{\fontfamily{cmr}\selectfont PTC}}      \\ \hline
		
		
		\hspace{-0.8em}	\textsc{DUGnn } &     $83.51$    	    &    $74.22$      \\  \hline
		\rowcolor[RGB]{252, 247, 246}  
		\hspace{-0.8em}	{\textsc{DUGnn} - {\fontfamily{cmr}\selectfont NCI1 / PTC} } &  $83.10$      &  $73.18$     \\    \hline
		
	\end{tabular}	
	\caption{  Ablation Study of Transfer Learning. \textsc{Dugnn} is the base model trained from scratch on both {\fontfamily{cmr}\selectfont NCI1 } and {\fontfamily{cmr}\selectfont   PTC} datasets via transfer learning. \textsc{Dugnn} - {\fontfamily{cmr}\selectfont NCI1 / PTC} represent  models trained from scratch on individual   datasets.   } 
	\label{table:transfer_perform}
	
\end{SCtable}

We conduct an ablation study to determine the importance of utilizing the pretrained  \textsc{Dugnn} model. In this experiment, we pick one of the cross validation splits of {\fontfamily{cmr}\selectfont NCI1 \& PTC} datasets for training \& validating, and fix all the hyper-parameters including the random seeds across the full ablation experiment.  We first train a \textsc{Dugnn} model on both {\fontfamily{cmr}\selectfont NCI1 \& PTC} datasets. Then the \textsc{Dugnn} models are trained on each dataset from scratch.  Table~\ref{table:transfer_perform} shows that  training without transfer learning reduces accuracy by around $\mathbf{0.4\textbf{\%}-1\textbf{\%}}$ on both datasets. Also, we see a bigger accuracy jump on {\fontfamily{cmr}\selectfont PTC}, since its dataset size is smaller, thus benefiting more via transfer learning.

\noindent \textbf{How much boost do we get with Multi-Task Graph Decoder?}

\renewcommand{\arraystretch}{2}
\begin{SCtable}[\sidecaptionrelwidth][h!]
	\centering
	\fontsize{7}{8}\selectfont
	\begin{tabular}{ p{3cm} |     K{1.6cm}  !{\vrule width0.8pt} K{1.6cm} !{\vrule width0.8pt} K{1.6cm}   | }
		
		\hspace{-1em}	\multirow{1}{*}{\textbf{Model / Dataset}} &       	\multicolumn{1}{c!{\vrule width0.8pt}}{PTC}  &  \multicolumn{1}{c!{\vrule width0.8pt}}{ENZYMES}      \\ \hline
		
		\rowcolor[RGB]{252, 247, 246}  	
		\hspace{-1em}	\textsc{DUGnn}   &    $73.53$  	    &    $65.00$  \\  \hline
		
		\hspace{-1em}	\textsc{DUGnn} - $\La_{A}$ - $\La_{\Ka}^{(\text{unsup})}$ &     $70.59$ 	    &  $61.67$    \\  \hline
		\rowcolor[RGB]{252, 247, 246}  
		\hspace{-1em}	\textsc{DUGnn} - $\La_{A}$     &  $72.68$   	   &  $64.10$  \\   \hline
		
		\hspace{-1em}	\textsc{DUGnn} - $\La_{\Ka}^{(\text{unsup})}$    &    $71.59$   	   &  $62.83$  \\   \hline
		\rowcolor[RGB]{252, 247, 246}  
		\hspace{-1em}		\textsc{DUGnn} -  $\La_{\text{class}}$  &  $64.71$      &  $56.53$    \\    \hline
		
		\Xhline{2\arrayrulewidth}
	\end{tabular}	
	\caption{\protect\rule{0ex}{4ex} Ablation Study of Multi-Task Decoder.	  \textsc{Dugnn} - $\ell_{(\cdot)}$ represents model trained from scratch without $\ell_{(\cdot)}$ loss function.  We pick one of the cross validation splits to report the accuracy and all the hyper-parameters, random seeds and data splits were kept constant across the ablation experiments.  } 
	\label{table:decoder_perform}
	
\end{SCtable}


In this ablation study,  we train \textsc{DUGnn} from scratch  with different loss functions.  Table~\ref{table:decoder_perform} reveals that completely removing the  graph decoder (i.e.,  $\La_{A}$ and $\La_{\Ka}^{(\text{unsup})}$) reduces accuracy by around $\mathbf{3\textbf{\%}-4\textbf{\%}}$ on both datasets. While removing only the graph kernel loss ($\La_{\Ka}^{(\text{unsup})}$) reduces accuracy by $\mathbf{2\textbf{\%}-3\textbf{\%}}$. The accuracy drops by around $\mathbf{1\textbf{\%}}$ when $\La_{A}$ is removed. Lastly, removing the supervised loss ($\La_{\text{class}}$)     reduces the performance considerably; nonetheless our model remains competitive and performs better against various graph kernel methods (see Table~\ref{table:bio_results}).

Ablation study related to effectiveness of adaptive supervised graph kernel loss is deferred to appendix.

%



\section{Conclusions}\label{sec:conclusion}

We have presented a   powerful {\em univeral graph embedding} neural network  architecture with three carefully designed components that can learn task-independent graph embeddings in a unsupervised fashion. In particular, the universal graph encoder component can be re-utilized across different datasets by leveraging transfer learning, and the  decoder component incorporates various  graph kernels to  capture   rich sub-structural properties to enable multi-task learning. 
Through extensive experiments and ablation studies on benchmark  graph classification datasets, we show that our proposed \textsc{DUGnn}  model can significantly outperform both the existing state-of-art graph neural network models and graph kernel methods. This demonstrates the  
the benefit of combining the power of graph neural networks in the design of a universal graph encoder with that of graph kernels in the design of a multi-task graph decoder. 

\setlength{\bibsep}{0pt}
\medskip
\small
\bibliographystyle{icml2019}
\bibliography{refs}

\begin{thebibliography}{58}
\providecommand{\natexlab}[1]{#1}
\providecommand{\url}[1]{\texttt{#1}}
\expandafter\ifx\csname urlstyle\endcsname\relax
  \providecommand{\doi}[1]{doi: #1}\else
  \providecommand{\doi}{doi: \begingroup \urlstyle{rm}\Url}\fi

\bibitem[Atwood \& Towsley(2016)Atwood and Towsley]{atwood2016diffusion}
Atwood, J. and Towsley, D.
\newblock Diffusion-convolutional neural networks.
\newblock In \emph{Advances in Neural Information Processing Systems}, pp.\
  1993--2001, 2016.

\bibitem[Belkin \& Niyogi(2003)Belkin and Niyogi]{belkin2003laplacian}
Belkin, M. and Niyogi, P.
\newblock Laplacian eigenmaps for dimensionality reduction and data
  representation.
\newblock \emph{Neural computation}, 15\penalty0 (6):\penalty0 1373--1396,
  2003.

\bibitem[Borgwardt \& Kriegel(2005)Borgwardt and
  Kriegel]{borgwardt2005shortest}
Borgwardt, K.~M. and Kriegel, H.-P.
\newblock Shortest-path kernels on graphs.
\newblock In \emph{Data Mining, Fifth IEEE International Conference on}, pp.\
  8--pp. IEEE, 2005.

\bibitem[Bousquet \& Elisseeff(2002)Bousquet and
  Elisseeff]{bousquet2002stability}
Bousquet, O. and Elisseeff, A.
\newblock Stability and generalization.
\newblock \emph{Journal of Machine Learning Research}, 2\penalty0
  (Mar):\penalty0 499--526, 2002.

\bibitem[Bruna et~al.(2013)Bruna, Zaremba, Szlam, and LeCun]{bruna2013spectral}
Bruna, J., Zaremba, W., Szlam, A., and LeCun, Y.
\newblock Spectral networks and locally connected networks on graphs.
\newblock \emph{arXiv preprint arXiv:1312.6203}, 2013.

\bibitem[Dai et~al.(2016)Dai, Dai, and Song]{dai2016discriminative}
Dai, H., Dai, B., and Song, L.
\newblock Discriminative embeddings of latent variable models for structured
  data.
\newblock In \emph{International Conference on Machine Learning}, pp.\
  2702--2711, 2016.

\bibitem[De~Cao \& Kipf(2018)De~Cao and Kipf]{de2018molgan}
De~Cao, N. and Kipf, T.
\newblock Molgan: An implicit generative model for small molecular graphs.
\newblock \emph{arXiv preprint arXiv:1805.11973}, 2018.

\bibitem[Defferrard et~al.(2016)Defferrard, Bresson, and
  Vandergheynst]{defferrard2016convolutional}
Defferrard, M., Bresson, X., and Vandergheynst, P.
\newblock Convolutional neural networks on graphs with fast localized spectral
  filtering.
\newblock In \emph{Advances in Neural Information Processing Systems}, pp.\
  3837--3845, 2016.

\bibitem[Dernbach et~al.(2018)Dernbach, Mohseni-Kabir, Pal, and
  Towsley]{dernbach2018quantum}
Dernbach, S., Mohseni-Kabir, A., Pal, S., and Towsley, D.
\newblock Quantum walk neural networks for graph-structured data.
\newblock In \emph{International Workshop on Complex Networks and their
  Applications}, pp.\  182--193. Springer, 2018.

\bibitem[Devlin et~al.(2018)Devlin, Chang, Lee, and Toutanova]{devlin2018bert}
Devlin, J., Chang, M.-W., Lee, K., and Toutanova, K.
\newblock Bert: Pre-training of deep bidirectional transformers for language
  understanding.
\newblock \emph{arXiv preprint arXiv:1810.04805}, 2018.

\bibitem[Duvenaud et~al.(2015)Duvenaud, Maclaurin, Iparraguirre, Bombarell,
  Hirzel, Aspuru-Guzik, and Adams]{duvenaud2015convolutional}
Duvenaud, D.~K., Maclaurin, D., Iparraguirre, J., Bombarell, R., Hirzel, T.,
  Aspuru-Guzik, A., and Adams, R.~P.
\newblock Convolutional networks on graphs for learning molecular fingerprints.
\newblock In \emph{Advances in neural information processing systems}, pp.\
  2224--2232, 2015.

\bibitem[Fey et~al.(2018)Fey, Lenssen, Weichert, and
  M{\"u}ller]{fey2018splinecnn}
Fey, M., Lenssen, J.~E., Weichert, F., and M{\"u}ller, H.
\newblock Splinecnn: Fast geometric deep learning with continuous b-spline
  kernels.
\newblock In \emph{Proceedings of the IEEE Conference on Computer Vision and
  Pattern Recognition}, pp.\  869--877, 2018.

\bibitem[Garc{\'\i}a-Dur{\'a}n \& Niepert(2017)Garc{\'\i}a-Dur{\'a}n and
  Niepert]{garcia2017learning}
Garc{\'\i}a-Dur{\'a}n, A. and Niepert, M.
\newblock Learning graph representations with embedding propagation.
\newblock \emph{arXiv preprint arXiv:1710.03059}, 2017.

\bibitem[G{\"a}rtner et~al.(2003)G{\"a}rtner, Flach, and
  Wrobel]{gartner2003graph}
G{\"a}rtner, T., Flach, P., and Wrobel, S.
\newblock On graph kernels: Hardness results and efficient alternatives.
\newblock In \emph{Learning Theory and Kernel Machines}, pp.\  129--143.
  Springer, 2003.

\bibitem[Gilmer et~al.(2017)Gilmer, Schoenholz, Riley, Vinyals, and
  Dahl]{gilmer2017neural}
Gilmer, J., Schoenholz, S.~S., Riley, P.~F., Vinyals, O., and Dahl, G.~E.
\newblock Neural message passing for quantum chemistry.
\newblock \emph{arXiv preprint arXiv:1704.01212}, 2017.

\bibitem[Hamilton et~al.(2017)Hamilton, Ying, and
  Leskovec]{hamilton2017inductive}
Hamilton, W., Ying, Z., and Leskovec, J.
\newblock Inductive representation learning on large graphs.
\newblock In \emph{Advances in Neural Information Processing Systems}, pp.\
  1024--1034, 2017.

\bibitem[Henaff et~al.(2015)Henaff, Bruna, and LeCun]{henaff2015deep}
Henaff, M., Bruna, J., and LeCun, Y.
\newblock Deep convolutional networks on graph-structured data.
\newblock \emph{arXiv preprint arXiv:1506.05163}, 2015.

\bibitem[Huang et~al.(2017)Huang, Liu, Van Der~Maaten, and
  Weinberger]{huang2017densely}
Huang, G., Liu, Z., Van Der~Maaten, L., and Weinberger, K.~Q.
\newblock Densely connected convolutional networks.
\newblock In \emph{CVPR}, volume~1, pp.\ ~3, 2017.

\bibitem[Ivanov \& Burnaev(2018)Ivanov and Burnaev]{ivanov2018anonymous}
Ivanov, S. and Burnaev, E.
\newblock Anonymous walk embeddings.
\newblock In \emph{Proceedings of the 35th International Conference on Machine
  Learning}, pp.\  2186--2195, 2018.

\bibitem[Karpathy et~al.(2014)Karpathy, Joulin, and Fei-Fei]{karpathy2014deep}
Karpathy, A., Joulin, A., and Fei-Fei, L.~F.
\newblock Deep fragment embeddings for bidirectional image sentence mapping.
\newblock In \emph{Advances in neural information processing systems}, pp.\
  1889--1897, 2014.

\bibitem[Kashima et~al.(2003)Kashima, Tsuda, and
  Inokuchi]{kashima2003marginalized}
Kashima, H., Tsuda, K., and Inokuchi, A.
\newblock Marginalized kernels between labeled graphs.
\newblock In \emph{ICML}, volume~3, pp.\  321--328, 2003.

\bibitem[Kiela \& Bottou(2014)Kiela and Bottou]{kiela2014learning}
Kiela, D. and Bottou, L.
\newblock Learning image embeddings using convolutional neural networks for
  improved multi-modal semantics.
\newblock In \emph{Proceedings of the 2014 Conference on Empirical Methods in
  Natural Language Processing (EMNLP)}, pp.\  36--45, 2014.

\bibitem[Kipf \& Welling(2016)Kipf and Welling]{kipf2016semi}
Kipf, T.~N. and Welling, M.
\newblock Semi-supervised classification with graph convolutional networks.
\newblock \emph{arXiv preprint arXiv:1609.02907}, 2016.

\bibitem[Kondor \& Pan(2016)Kondor and Pan]{kondor2016multiscale}
Kondor, R. and Pan, H.
\newblock The multiscale laplacian graph kernel.
\newblock In \emph{Advances in Neural Information Processing Systems}, pp.\
  2982--2990, 2016.

\bibitem[Kondor et~al.(2009)Kondor, Shervashidze, and
  Borgwardt]{kondor2009graphlet}
Kondor, R., Shervashidze, N., and Borgwardt, K.~M.
\newblock The graphlet spectrum.
\newblock In \emph{Proceedings of the 26th Annual International Conference on
  Machine Learning}, pp.\  529--536. ACM, 2009.

\bibitem[Lei et~al.(2017)Lei, Jin, Barzilay, and Jaakkola]{lei2017deriving}
Lei, T., Jin, W., Barzilay, R., and Jaakkola, T.
\newblock Deriving neural architectures from sequence and graph kernels.
\newblock \emph{arXiv preprint arXiv:1705.09037}, 2017.

\bibitem[Levie et~al.(2017)Levie, Monti, Bresson, and
  Bronstein]{levie2017cayleynets}
Levie, R., Monti, F., Bresson, X., and Bronstein, M.~M.
\newblock Cayleynets: Graph convolutional neural networks with complex rational
  spectral filters.
\newblock \emph{arXiv preprint arXiv:1705.07664}, 2017.

\bibitem[Li et~al.(2018{\natexlab{a}})Li, Han, and Wu]{li2018deeper}
Li, Q., Han, Z., and Wu, X.-M.
\newblock Deeper insights into graph convolutional networks for semi-supervised
  learning.
\newblock \emph{arXiv preprint arXiv:1801.07606}, 2018{\natexlab{a}}.

\bibitem[Li et~al.(2018{\natexlab{b}})Li, Wang, Zhu, and Huang]{li2018adaptive}
Li, R., Wang, S., Zhu, F., and Huang, J.
\newblock Adaptive graph convolutional neural networks.
\newblock \emph{arXiv preprint arXiv:1801.03226}, 2018{\natexlab{b}}.

\bibitem[McCann et~al.(2018)McCann, Keskar, Xiong, and
  Socher]{mccann2018natural}
McCann, B., Keskar, N.~S., Xiong, C., and Socher, R.
\newblock The natural language decathlon: Multitask learning as question
  answering.
\newblock \emph{arXiv preprint arXiv:1806.08730}, 2018.

\bibitem[Mikolov et~al.(2013)Mikolov, Sutskever, Chen, Corrado, and
  Dean]{mikolov2013distributed}
Mikolov, T., Sutskever, I., Chen, K., Corrado, G.~S., and Dean, J.
\newblock Distributed representations of words and phrases and their
  compositionality.
\newblock In \emph{Advances in neural information processing systems}, pp.\
  3111--3119, 2013.

\bibitem[Niepert et~al.(2016)Niepert, Ahmed, and Kutzkov]{niepert2016learning}
Niepert, M., Ahmed, M., and Kutzkov, K.
\newblock Learning convolutional neural networks for graphs.
\newblock In \emph{Proceedings of the 33rd annual international conference on
  machine learning. ACM}, 2016.

\bibitem[Orsini et~al.(2015)Orsini, Frasconi, and De~Raedt]{orsini2015graph}
Orsini, F., Frasconi, P., and De~Raedt, L.
\newblock Graph invariant kernels.
\newblock In \emph{IJCAI}, pp.\  3756--3762, 2015.

\bibitem[Paratte \& Martin(2016)Paratte and Martin]{paratte2016fast}
Paratte, J. and Martin, L.
\newblock Fast eigenspace approximation using random signals.
\newblock \emph{arXiv preprint arXiv:1611.00938}, 2016.

\bibitem[Pennington et~al.(2014)Pennington, Socher, and
  Manning]{pennington2014glove}
Pennington, J., Socher, R., and Manning, C.
\newblock Glove: Global vectors for word representation.
\newblock In \emph{Proceedings of the 2014 conference on empirical methods in
  natural language processing (EMNLP)}, pp.\  1532--1543, 2014.

\bibitem[Peters et~al.(2018)Peters, Neumann, Iyyer, Gardner, Clark, Lee, and
  Zettlemoyer]{peters2018deep}
Peters, M.~E., Neumann, M., Iyyer, M., Gardner, M., Clark, C., Lee, K., and
  Zettlemoyer, L.
\newblock Deep contextualized word representations.
\newblock \emph{arXiv preprint arXiv:1802.05365}, 2018.

\bibitem[Pr{\v{z}}ulj(2007)]{prvzulj2007biological}
Pr{\v{z}}ulj, N.
\newblock Biological network comparison using graphlet degree distribution.
\newblock \emph{Bioinformatics}, 23\penalty0 (2):\penalty0 e177--e183, 2007.

\bibitem[Puy et~al.(2017)Puy, Kitic, and P{\'e}rez]{puy2017unifying}
Puy, G., Kitic, S., and P{\'e}rez, P.
\newblock Unifying local and non-local signal processing with graph cnns.
\newblock \emph{arXiv preprint arXiv:1702.07759}, 2017.

\bibitem[Sch{\"u}tt et~al.(2017)Sch{\"u}tt, Arbabzadah, Chmiela, M{\"u}ller,
  and Tkatchenko]{schutt2017quantum}
Sch{\"u}tt, K.~T., Arbabzadah, F., Chmiela, S., M{\"u}ller, K.~R., and
  Tkatchenko, A.
\newblock Quantum-chemical insights from deep tensor neural networks.
\newblock \emph{Nature communications}, 8:\penalty0 13890, 2017.

\bibitem[Shervashidze et~al.(2009)Shervashidze, Vishwanathan, Petri, Mehlhorn,
  and Borgwardt]{shervashidze2009efficient}
Shervashidze, N., Vishwanathan, S., Petri, T., Mehlhorn, K., and Borgwardt,
  K.~M.
\newblock Efficient graphlet kernels for large graph comparison.
\newblock In \emph{AISTATS}, volume~5, pp.\  488--495, 2009.

\bibitem[Shervashidze et~al.(2011)Shervashidze, Schweitzer, Leeuwen, Mehlhorn,
  and Borgwardt]{shervashidze2011weisfeiler}
Shervashidze, N., Schweitzer, P., Leeuwen, E. J.~v., Mehlhorn, K., and
  Borgwardt, K.~M.
\newblock Weisfeiler-lehman graph kernels.
\newblock \emph{Journal of Machine Learning Research}, 12\penalty0
  (Sep):\penalty0 2539--2561, 2011.

\bibitem[Shuman et~al.(2013)Shuman, Narang, Frossard, Ortega, and
  Vandergheynst]{shuman2013emerging}
Shuman, D.~I., Narang, S.~K., Frossard, P., Ortega, A., and Vandergheynst, P.
\newblock The emerging field of signal processing on graphs: Extending
  high-dimensional data analysis to networks and other irregular domains.
\newblock \emph{IEEE Signal Processing Magazine}, 30\penalty0 (3):\penalty0
  83--98, 2013.

\bibitem[Simonovsky \& Komodakis(2017)Simonovsky and
  Komodakis]{simonovsky2017dynamic}
Simonovsky, M. and Komodakis, N.
\newblock Dynamic edge-conditioned filters in convolutional neural networks on
  graphs.
\newblock In \emph{Proc. CVPR}, 2017.

\bibitem[Simonovsky \& Komodakis(2018)Simonovsky and
  Komodakis]{simonovsky2018graphvae}
Simonovsky, M. and Komodakis, N.
\newblock Graphvae: Towards generation of small graphs using variational
  autoencoders.
\newblock \emph{arXiv preprint arXiv:1802.03480}, 2018.

\bibitem[Vaswani et~al.(2017)Vaswani, Shazeer, Parmar, Uszkoreit, Jones, Gomez,
  Kaiser, and Polosukhin]{vaswani2017attention}
Vaswani, A., Shazeer, N., Parmar, N., Uszkoreit, J., Jones, L., Gomez, A.~N.,
  Kaiser, {\L}., and Polosukhin, I.
\newblock Attention is all you need.
\newblock In \emph{Advances in Neural Information Processing Systems}, pp.\
  5998--6008, 2017.

\bibitem[Veli{\v{c}}kovi{\'c} et~al.(2018)Veli{\v{c}}kovi{\'c}, Fedus,
  Hamilton, Li{\`o}, Bengio, and Hjelm]{velivckovic2018deep}
Veli{\v{c}}kovi{\'c}, P., Fedus, W., Hamilton, W.~L., Li{\`o}, P., Bengio, Y.,
  and Hjelm, R.~D.
\newblock Deep graph infomax.
\newblock \emph{arXiv preprint arXiv:1809.10341}, 2018.

\bibitem[Verma \& Zhang(2017)Verma and Zhang]{verma2017hunt}
Verma, S. and Zhang, Z.-L.
\newblock Hunt for the unique, stable, sparse and fast feature learning on
  graphs.
\newblock In \emph{Advances in Neural Information Processing Systems}, pp.\
  87--97, 2017.

\bibitem[Verma \& Zhang(2018)Verma and Zhang]{verma2018graph}
Verma, S. and Zhang, Z.-L.
\newblock Graph capsule convolutional neural networks.
\newblock \emph{arXiv preprint arXiv:1805.08090}, 2018.

\bibitem[Verma \& Zhang(2019)Verma and Zhang]{verma2019stability}
Verma, S. and Zhang, Z.-L.
\newblock Stability and generalization of graph convolutional neural networks.
\newblock \emph{arXiv preprint arXiv:1905.01004}, 2019.

\bibitem[Wu et~al.(2018)Wu, Ramsundar, Feinberg, Gomes, Geniesse, Pappu,
  Leswing, and Pande]{wu2018moleculenet}
Wu, Z., Ramsundar, B., Feinberg, E.~N., Gomes, J., Geniesse, C., Pappu, A.~S.,
  Leswing, K., and Pande, V.
\newblock Moleculenet: a benchmark for molecular machine learning.
\newblock \emph{Chemical science}, 9\penalty0 (2):\penalty0 513--530, 2018.

\bibitem[Xu et~al.(2018)Xu, Hu, Leskovec, and Jegelka]{xu2018powerful}
Xu, K., Hu, W., Leskovec, J., and Jegelka, S.
\newblock How powerful are graph neural networks?
\newblock \emph{arXiv preprint arXiv:1810.00826}, 2018.

\bibitem[Yanardag \& Vishwanathan(2015)Yanardag and
  Vishwanathan]{yanardag2015deep}
Yanardag, P. and Vishwanathan, S.
\newblock Deep graph kernels.
\newblock In \emph{Proceedings of the 21th ACM SIGKDD International Conference
  on Knowledge Discovery and Data Mining}, pp.\  1365--1374. ACM, 2015.

\bibitem[Ying et~al.(2018)Ying, You, Morris, Ren, Hamilton, and
  Leskovec]{ying2018hierarchical}
Ying, Z., You, J., Morris, C., Ren, X., Hamilton, W., and Leskovec, J.
\newblock Hierarchical graph representation learning with differentiable
  pooling.
\newblock In \emph{Advances in Neural Information Processing Systems}, pp.\
  4805--4815, 2018.

\bibitem[You et~al.(2018{\natexlab{a}})You, Liu, Ying, Pande, and
  Leskovec]{you2018graph}
You, J., Liu, B., Ying, R., Pande, V., and Leskovec, J.
\newblock Graph convolutional policy network for goal-directed molecular graph
  generation.
\newblock \emph{arXiv preprint arXiv:1806.02473}, 2018{\natexlab{a}}.

\bibitem[You et~al.(2018{\natexlab{b}})You, Ying, Ren, Hamilton, and
  Leskovec]{you2018graphrnn}
You, J., Ying, R., Ren, X., Hamilton, W., and Leskovec, J.
\newblock Graphrnn: Generating realistic graphs with deep auto-regressive
  models.
\newblock In \emph{International Conference on Machine Learning}, pp.\
  5694--5703, 2018{\natexlab{b}}.

\bibitem[Zhang et~al.(2016)Zhang, Yuan, Lian, Xie, and
  Ma]{zhang2016collaborative}
Zhang, F., Yuan, N.~J., Lian, D., Xie, X., and Ma, W.-Y.
\newblock Collaborative knowledge base embedding for recommender systems.
\newblock In \emph{Proceedings of the 22nd ACM SIGKDD international conference
  on knowledge discovery and data mining}, pp.\  353--362. ACM, 2016.

\bibitem[Zhang et~al.(2018)Zhang, Cui, Neumann, and Chen]{zhang2018end}
Zhang, M., Cui, Z., Neumann, M., and Chen, Y.
\newblock An end-to-end deep learning architecture for graph classification.
\newblock In \emph{AAAI}, pp.\  4438--4445, 2018.

\bibitem[Zhang et~al.(2019)Zhang, Chen, Wang, Bai, and
  Hancock]{zhang2019quantum}
Zhang, Z., Chen, D., Wang, J., Bai, L., and Hancock, E.~R.
\newblock Quantum-based subgraph convolutional neural networks.
\newblock \emph{Pattern Recognition}, 88:\penalty0 38--49, 2019.

\end{thebibliography}

\section{Appendix}

\section{Proof of Theorem 1}


The proof of Theorem 1 follows immediately from a key result in ~\cite{paratte2016fast}  stated below.

\begin{lemma}~\cite{paratte2016fast} \label{lemma:projection} 
	Let $\mathbf{U} \in \mathbb{R}^{N\times N}$ be an orthonormal matrix and $\mathbf{R} \in \mathbb{R}^{N\times d}$  be a Gaussian random matrix with i.i.d. entries $\sim \Na(0, \sigma^2)$. Then the entries of $\mathbf{U} \mathbf{R}$ are i.i.d. Gaussian random samples with the same pdf  $\sim \Na(0, \sigma^2)$.
\end{lemma}

Let $f(\mathbf{L})$ be a function of graph Laplacian that operates on eigenvalues and defined as follows $f(\mathbf{L})=\mathbf{U}f(\mathbf{\Sigma})\mathbf{U}^{T}$  where $\mathbf{\Sigma}$ is the eigenvalue diagonal matrix  and $\mathbf{U}$ is the eigenvector matrix which is also an orthonormal matrix. Therefore, $f(\mathbf{L})  \mathbf{R}=\mathbf{U}f(\mathbf{\Sigma})\mathbf{U}^{T}  \mathbf{R} =\mathbf{U}f(\mathbf{\Sigma})  \mathbf{\widehat{R}} $ where $\mathbf{\widehat{R}} \in \mathbb{R}^{N\times d}$ is again a  Gaussian random matrix from Lemma~\ref{lemma:projection}. Here $\mathbf{\widehat{U}} = \mathbf{U}f(\mathbf{\Sigma})$ contains the eigenvector columns  scaled by the $f(\mathbf{\Sigma})$ eigenvalues. Thus $\mathbf{\widehat{U}} \mathbf{\widehat{R}}$ results in a random projection  (or dimension reduction) of the
scaled eigenvector space of the graph Laplacian into a $d$-dimensional space. 

\section{Proof of Theorem 2}

Theorem 2's subpart where \textsc{DUGnn} representation power is \emph{same} as  Weisfeiler-Lehman (WL) graph isomorphism test  directly follows from  Theorem 3 of the paper~\cite{xu2018powerful}. Note that both the conditions  stated in Theorem 3 are met,  since our node aggregate function and readout function is same as  Graph Isomorphism Network (GIN). To show that \textsc{DUGnn} is \emph{atleast} as powerful than WL, we provide the general condition on $k$-regular graphs to be hold such that their graph representations are different. It is suffice   to show  a single example where WL fails but \textsc{DUGnn} succeeds.

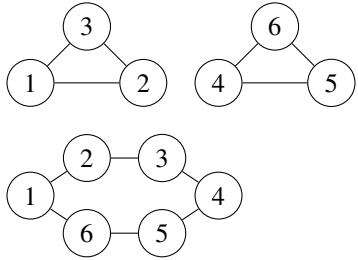
\begin{figure}[!h]
	\centering
	\begin{tikzpicture}
	\node[draw, circle] at (0, 0)   (n1) {1};
	\node[draw, circle] at (3, 0)   (n2) {2};
	\node[draw, circle] at (1.5, 1.5)   (n3) {3};
	
	\node[draw, circle] at (5, 0)   (n4) {4};
	\node[draw, circle] at (8, 0)   (n5) {5};
	\node[draw, circle] at (6.5, 1.5) (n6) {6};
	
	\node[draw, circle] at (0, -3)   (m1) {1};
	\node[draw, circle] at (1.5, -2)   (m2) {2};
	\node[draw, circle] at (3.5, -2)   (m3) {3};
	\node[draw, circle] at (5, -3)   (m4) {4};
	\node[draw, circle] at (3.5, -4)   (m5) {5};
	\node[draw, circle] at (1.5, -4)   (m6) {6};
	\draw[] (n1) -- (n2);
	\draw[] (n2) -- (n3);
	\draw[] (n3) -- (n1);
	\draw[] (n4) -- (n5);
	\draw[] (n5) -- (n6);
	\draw[] (n6) -- (n4);
	
	\draw[] (m1) -- (m2);
	\draw[] (m2) -- (m3);
	\draw[] (m3) -- (m4);
	\draw[] (m4) -- (m5);
	\draw[] (m5) -- (m6);
	\draw[] (m6) -- (m1);
	
	\end{tikzpicture}
	\caption{Two Regular Graphs $G_1$ and $G_2$.}\label{fig:reg_graphs}
\end{figure}

Consider two $k$-regular non-isomorphic graphs of size $N$ and let $\Na(i)$ represents the neighbor indices of $i^{th}$ node including itself.  Since node features initialized are identical (or equal to node degree) in classic WL  i.e, $x_v^{(0)} = c$ for each node $v$, after local aggregation operation, the new node features is given by $x_v^{(1)}=\phi(c, f(\underset{\text{k-times}}{\{c,...,c\}}))$ for  each node in $G_1$ where $f$ is a multi-set function and $\phi$ is a injective function. Similarly for $G_2$ each node representation is given by $x_v^{(1)}=\phi(c, f(\underset{\text{k-times}}{\{c,...,c\}}))$ after first iteration which is same as $G_1$. As a result, the graph representation of $G_1$ and $G_2$ are same since sum-pool operation yields the same value i.e., $\sum_{v}x_v^{(1)}$ in both cases. 

Consider \textsc{DUGnn}  initialized  with graph spectral embeddings  as node features. Let $\{u_1,u_2, ..., u_N\}$ and $\{v_1, v_2, ..., v_N\}$ be the spectral embeddings of nodes in $G_1$and $G_2$. Then, updated representation  of   $i^{th}$ node is given by  $x_i^{(1)}=\phi(\sum_{j \in \Na(i)}  u_{j})$ where $\phi$ is the universal MLP function.  Let $\phi$ MLP  function learns an identity function. Then the sum-pool representation of $G_1$ is given by $y_1  = \sum_{i=1}^{N}\sum_{j \in \Na(i)}  u_{j} = k \sum_{i=1}^{N} u_i$. Similarly, for $G_2$ the final representation is  $y_2= k\sum_{i=1}^{N} v_i$.  As a result, for all non-isomorphic $k$-regular graph pairs whose  row-wise sum of eigenvector matrix respectively are not equal -- can be distinguished by \textsc{DUGnn} -- but not possible in the case of   classic WL.  It is easy to verify numerically that the row-wise sum of eigenvector matrices  of graphs   shown in Figure~\ref{fig:reg_graphs}  are not equal which implies  $y_1\neq y_2$ and hence their graph representations  are different.

\section{Proof of Theorem 3}

Theorem 3 is based on main result of the paper~\cite{verma2019stability} and the extension to graph classification as well as graph capsule networks. We  show that the single layer graph capsule networks are also uniformly stable~\cite{bousquet2002stability} for solving  graph  classification problem. Like in ~\cite{verma2019stability}, we assume that the activation and loss functions are Lipschitz continuous and smooth. For convenience, we borrow the same notions used in~\cite{verma2019stability}   and  similarly divide the proofs in two parts. In first part, we separate out the  terms due to  weight parameters and graph convolution operation in order to bound their difference   independently. In second part, we bound the expected difference (due to   to randomness of SGD algorithm) in weight parameters  under single data perturbation. 

Following is a single layer Universal Graph Encoder  GNN function $f(\mathbf{x},\bm{\uptheta} ) \in \mathbb{R}$   based on statistical moments with sum as read-out function in graph classification task where $N$ is number of nodes in a graph, $\mathbf{x}_i$ is the $i^{th}$ node feature value and $\bm{\uptheta}$ is the capsule learning parameter. For simplicity, we show our results for the case of  $\mathbf{x} \in \mathbb{R}$ but the proof remains applicable for general  $\mathbf{x} \in \mathbb{R}^{d}$.
\begin{equation}\label{eq:gcnn_node}
\begin{split}
f(\mathbf{x}=\{\mathbf{x}_1, \dots, \mathbf{x}_N \},\bm{\uptheta} ) &= \sum_{n=1}^{N} \Bigg(\sigma\Big( \sum_{\mathclap{\substack{j\in \\ \Na(\mathbf{x}_n)}}}  e_{\cdot j}  \mathbf{x}_{j}  \bm{\uptheta}_{1} \Big) + \sigma\Big( \sum_{\mathclap{\substack{j\in \\ \Na(\mathbf{x}_n)}}}  e_{\cdot j}  \mathbf{x}^{2}_{j}  \bm{\uptheta}_{2} \Big) + \cdots + \sigma\Big( \sum_{\mathclap{\substack{j\in \\ \Na(\mathbf{x}_n)}}}  e_{\cdot j}  \mathbf{x}^{2}_{j}  \bm{\uptheta}_{p} \Big) \Bigg)  \\
&=  \sum_{n=1}^{N} \Bigg( \sum_{p=1}^{P}\sigma \Big( \sum_{\mathclap{\substack{j\in \\ \Na(\mathbf{x}_n)}}}  e_{\cdot j}  \mathbf{x}^{p}_{j}  \bm{\uptheta}_{p} \Big) \Bigg)   \\
\end{split}
\end{equation}
The first order derivative with respect to $p^{th}-$parameter is given as,
\begin{equation}~\label{eq:gcnn_derivative}	
\frac{\partial f(\mathbf{x},\bm{\uptheta} )}{\partial \bm{\uptheta}_{p} }  = \sum_{n=1}^{N} \Bigg( \sigma' \Big( \sum_{\mathclap{\substack{j\in \\ \Na(\mathbf{x})}}}  e_{\cdot j}  \mathbf{x}^{p}_{j}  \bm{\uptheta}_{p} \Big)  \sum_{\mathclap{\substack{j\in \\ \Na(\mathbf{x}) }}}  e_{\cdot j}  \mathbf{x}^{p}_{j} \Bigg)   
\end{equation}

where $P$ is number of instantiation parameters.

\noindent \textbf{Proof Part I}:   Let $\mathbf{E}_{\textsc{sgd}}$ be expectation due to SGD randomness and let $\bm{\uptheta}_S$ and  $\bm{\uptheta}_{S^i}$ represent the     filter weights learned on training set $S$ and $S^i$ that differs in precisely single data point. Also,   $\Delta \bm{\uptheta} = \bm{\uptheta}_S  - \bm{\uptheta}_{S^i}  $. 

\begin{equation}~\label{eq:proof_part1}	
\begin{split}
& |\mathbf{E}_{\textsc{sgd}}[\ell(A_S,y) - \ell(A_{S^{i}} ,y)]|  \leq \alpha_{\ell} \mathbf{E}_{\textsc{sgd}}[| f(\mathbf{x},\bm{\uptheta}_S) -  f(\mathbf{x},\bm{\uptheta}_{S^{i}})|] \\
&  \leq  \alpha_{\ell} \mathbf{E}_{\textsc{sgd}}\Big[\Big|  \sum_{n=1}^{N} \Bigg( \sum_{p=1}^{P}  \sigma\Big(\sum_{\mathclap{\substack{j\in \\ \Na(\mathbf{x}_n)}}} e_{\cdot j} \mathbf{x}^{p}_{j} \bm{\uptheta}_{p,S}  \Big)  \Bigg) - \sum_{n=1}^{N} \Bigg( \sum_{p=1}^{P}\sigma\Big( \sum_{\mathclap{\substack{j \in \\ \Na(\mathbf{x}_n)}}} \  e_{\cdot j}   \mathbf{x}^{p}_{j} \bm{\uptheta}_{p, S^{i}} \Big)  \Bigg) \Big|\Big]  \\
&  \leq  \alpha_{\ell} \mathbf{E}_{\textsc{sgd}}\Big[  \sum_{n=1}^{N} \sum_{p=1}^{P}\Big|   \sigma\Big(\sum_{\mathclap{\substack{j\in \\ \Na(\mathbf{x}_n)}}} e_{\cdot j} \mathbf{x}^{p}_{j} \bm{\uptheta}_{p,S}  \Big) -  \sigma\Big( \sum_{\mathclap{\substack{j \in \\ \Na(\mathbf{x}_n)}}} \  e_{\cdot j}   \mathbf{x}^{p}_{j} \bm{\uptheta}_{p, S^{i}} \Big) \Big|\Big]   \\
\end{split}
\end{equation}
\begin{equation*} 
\begin{split}
& \text{Since activation function is also $\sigma-$Lipschitz  continuous,} \\
&   \leq \alpha_{\ell} \mathbf{E}_{\textsc{sgd}}\Big[  \sum_{n=1}^{N} \sum_{p=1}^{P} \Big|    \sum_{\mathclap{\substack{j\in \\ \Na(\mathbf{x}_n)}}} e_{\cdot j} \mathbf{x}^{p}_{j} \bm{\uptheta}_{p,S}    -   \sum_{\mathclap{\substack{j \in \\ \Na(\mathbf{x}_n)}}} \  e_{\cdot j}   \mathbf{x}^{p}_{j} \bm{\uptheta}_{p, S^{i}}  \Big|\Big]   \\
& \leq \alpha_{\ell} \mathbf{E}_{\textsc{sgd}}\Big[ \sum_{n=1}^{N}\sum_{p=1}^{P} \Big|   \sum_{\mathclap{\substack{j\in \\ \Na(\mathbf{x}_n)}}} e_{\cdot j} \mathbf{x}^{p}_{j} (\bm{\uptheta}_{p, S}    -  \bm{\uptheta}_{p, S^{i}} )  \Big|\Big]   \\
& \leq \alpha_{\ell} \mathbf{E}_{\textsc{sgd}}\Big[ \sum_{n=1}^{N}\sum_{p=1}^{P} \Bigg( \big|\sum_{\mathclap{\substack{j \in \\ \Na(\mathbf{x}_n)}}} \big( e_{\cdot j }  \mathbf{x}^{p}_{j}  \big)\big| \big(\big|\bm{\uptheta}_{p,S}  - \bm{\uptheta}_{p,S^{i}} \big|\big) \Bigg) \Big]   \\
& \leq \alpha_{\ell} \sum_{n=1}^{N} \sum_{p=1}^{P} \Bigg( \big| \sum_{\mathclap{\substack{j \in \\ \Na(\mathbf{x}_n)}}} \big( e_{\cdot j }  \mathbf{x}^{p}_{j}  \big)\big| \big( \mathbf{E}_{\textsc{sgd}}\big[\big|\Delta \bm{\uptheta}_{p} \big|\big]   \big)    \Bigg)  \\
& \leq \alpha_{\ell} N\sum_{p=1}^{P} \Big( \mathbf{g}_{p,\lambda}   \mathbf{E}_{\textsc{sgd}}\big[\big|\Delta \bm{\uptheta}_{p} \big|\big]  \Big)    \\
\end{split}
\end{equation*}

where  
$\mathbf{g}_{p,\lambda}$ is defined as $  \mathbf{g}_{p,\lambda}:=\underset{\mathbf{x}}{\sup}  \hspace{0.5em} \Big|\sum_{j \in  \Na(\mathbf{x})}   e_{\cdot j }  \mathbf{x}^{p}_{j}  \Big| $.
The term $\mathbf{g}_{p,\lambda}$ will later be bounded in terms of the largest absolute eigenvalue of the graph convolution filter $g(\mathbf{L})$.  

\noindent \textbf{Proof Part II}:   We move on the second part of the proof where we bound the different in weights learned due to single data perturbation in Lemma~\ref{lemma:sgd_terms} using Lemma~\ref{lemma:sgd_term1} and Lemma~\ref{lemma:sgd_term2}.

\begin{lemma}[\textbf{Universal Graph Encoder    Same Sample Loss Stability Bound}]\label{lemma:sgd_term1}  \textit{The   loss-derivative bound difference   of (single-layer) Universal Graph Encoder    trained with SGD algorithm for $T$ iterations on two training datasets $S$ and $S^i$ respectively, with respect to the same sample     is given by,}
	$$ \Big|   \nabla_{p} \ell\big(f(\mathbf{x},\bm{\uptheta}_{S, t} ),y \big)   -      \nabla_{p} \ell\big(f(\mathbf{x},\bm{\uptheta}_{S^{i}, t} ),y \big)   \Big| \leq \nu_{\ell}  \nu_{\sigma} N \mathbf{g}_{p, \lambda}^2  |   \Delta\bm{\uptheta}_{p, t}|  $$
\end{lemma}
\noindent \textbf{Proof}:  Using Equation~\ref{eq:gcnn_derivative} here, we get,	

\begin{equation} 	
\begin{split}
& \Big|   \frac{\partial \ell\big(f(\mathbf{x},\bm{\uptheta}_{S, t} ),y \big)}{\partial \bm{\uptheta}_{p} }    -      \frac{\partial \ell \big(f(\mathbf{x},\bm{\uptheta}_{S^{i}, t} ),y \big)}{\partial \bm{\uptheta}_{p} }     \Big| \nonumber \leq   \nu_{\ell} \Big| \frac{\partial f(\mathbf{x},\bm{\uptheta}_{S, t} )}{\partial \bm{\uptheta}_{p} }     -  \frac{\partial f(\mathbf{x},\bm{\uptheta}_{S^{i}, t} )}{\partial \bm{\uptheta}_{p} }   \Big|  \\
& \leq \nu_{\ell} \Big|  \sum_{n=1}^{N} \Bigg( \sigma^{'}\Big( \sum_{\mathclap{\substack{j\in \\ \Na(\mathbf{x}_n)}}}  e_{\cdot j}  \mathbf{x}^{p}_{j}  \bm{\uptheta}_{p, S, t} \Big)  \sum_{\mathclap{\substack{j\in \\ \Na(\mathbf{x}_n) }}}  e_{\cdot j}  \mathbf{x}^{p}_{j}  \Bigg) -  \sum_{n=1}^{N}  \Bigg( \sigma^{'}\Big( \sum_{\mathclap{\substack{j\in \\ \Na(\mathbf{x}_n)}}}  e_{\cdot j}  \mathbf{x}^{p}_{j}  \bm{\uptheta}_{p, S^{i}, t} \Big)  \sum_{\mathclap{\substack{j\in \\ \Na(\mathbf{x}_n) }}}  e_{\cdot j}  \mathbf{x}^{p}_{j}   \Bigg) \Big|  \\
& \leq \nu_{\ell} \sum_{n=1}^{N}  \Bigg( \Big( \big| \sum_{\mathclap{\substack{j\in \\ \Na(\mathbf{x}_n) }}}  e_{\cdot j}  \mathbf{x}^{p}_{j} \big|  \Big) \Big|   \sigma^{'}\Big( \sum_{\mathclap{\substack{j\in \\ \Na(\mathbf{x}_n)}}}  e_{\cdot j}  \mathbf{x}^{p}_{j}  \bm{\uptheta}_{p, S, t} \Big)   -    \sigma^{'}\Big( \sum_{\mathclap{\substack{j\in \\ \Na(\mathbf{x}_n)}}}  e_{\cdot j}  \mathbf{x}^{p}_{j}  \bm{\uptheta}_{p, S^{i}, t} \Big)    \Big| \Bigg)  \\
& \text{Since the activation function is    Lipschitz continuous and smooth,  } \\
& \text{and plugging  $\big| \sum_{\mathclap{\substack{j\in \\ \Na(\mathbf{x}_n) }}}  e_{\cdot j}  \mathbf{x}^{p}_{j} \big| \leq \mathbf{g}_{p,\lambda}$, we get,  } \\
\end{split}
\end{equation}
\begin{equation*} 	
\begin{split}
& \leq \nu_{\ell}  \nu_{\sigma}   \sum_{n=1}^{N}  \Bigg( \mathbf{g}_{p,\lambda}   \Big|  \Big( \sum_{\mathclap{\substack{j\in \\ \Na(\mathbf{x}_n)}}}  e_{\cdot j}  \mathbf{x}^{p}_{j}  \bm{\uptheta}_{p, S, t} \Big)   -  \Big( \sum_{\mathclap{\substack{j\in \\ \Na(\mathbf{x}_n)}}}  e_{\cdot j}  \mathbf{x}^{p}_{j}  \bm{\uptheta}_{p, S^{i}, t} \Big)    \Big|   \Bigg) \\
& \leq \nu_{\ell}  \nu_{\sigma} \sum_{n=1}^{N}  \Bigg(  \mathbf{g}_{p, \lambda} \Big( \big| \sum_{\mathclap{\substack{j\in \\ \Na(\mathbf{x}_n)}}}  e_{\cdot j}  \mathbf{x}^{p}_{j}  \big| \Big)   |\bm{\uptheta}_{p, S, t} - \bm{\uptheta}_{p, S^{i}, t}| \Bigg) \\
& \leq \nu_{\ell}  \nu_{\sigma} N \mathbf{g}_{p, \lambda}^2 | \Delta\bm{\uptheta}_{p, t}|  
\end{split}
\end{equation*}
This completes the proof of Lemma~\ref{lemma:sgd_term1}. \\

\begin{lemma}[\textbf{Universal Graph Encoder    Different Sample Loss Stability Bound}]\label{lemma:sgd_term2}
	\textit{The   loss-derivative bound difference   of (single-layer) Universal Graph Encoder     trained with SGD algorithm for $T$ iterations on two training datasets $S$ and $S^i$ respectively,  with respect to  the  different samples     is given by,}
	$$ \Big|   \nabla_{p} \ell\big(f(\mathbf{x}_i,\bm{\uptheta}_{S, t} ),y_i \big)   -      \nabla_{p} \ell\big(f(\mathbf{x}'_i,\bm{\uptheta}_{S^{i}, t} ),y'_i \big)   \Big| \leq 2\nu_{\ell} \alpha_{\sigma} N\mathbf{g}_{p, \lambda}. $$
\end{lemma}
\noindent \textbf{Proof}:

\begin{equation} 	
\begin{split}
& \Big|   \frac{\partial \ell\big(f(\mathbf{x},\bm{\uptheta}_{S, t} ),y \big)}{\partial \bm{\uptheta}_{p} }    -      \frac{\partial \ell \big(f(\mathbf{\tilde{x}} ,\bm{\uptheta}_{S^{i}, t} ),\tilde{y} \big)}{\partial \bm{\uptheta}_{p} }     \Big| \nonumber \leq   \nu_{\ell} \Big| \frac{\partial f(\mathbf{x},\bm{\uptheta}_{S, t} )}{\partial \bm{\uptheta}_{p} }     -  \frac{\partial f(\mathbf{\tilde{x}},\bm{\uptheta}_{S^{i}, t} )}{\partial \bm{\uptheta}_{p} }   \Big|  \\
& \leq \nu_{\ell} \Big| \sum_{n=1}^{N}  \Bigg(  \sigma^{'} \Big( \sum_{\mathclap{\substack{j\in \\ \Na(\mathbf{x}_n)}}}  e_{\cdot j}  \mathbf{x}^{p}_{j}  \bm{\uptheta}_{p, S, t} \Big)  \sum_{\mathclap{\substack{j\in \\ \Na(\mathbf{x}_n) }}}  e_{\cdot j}  \mathbf{x}^{p}_{j} \Bigg) -    \sum_{n=1}^{N}  \Bigg( \sigma^{'} \Big( \sum_{\mathclap{\substack{j\in \\ \Na(\mathbf{\tilde{x}}_n )}}}  e_{\cdot j}  \mathbf{\tilde{x}}^{p}_{j}   \bm{\uptheta}_{p, S^{i}, t} \Big)  \sum_{\mathclap{\substack{j\in \\ \Na(\mathbf{\tilde{x}}_n ) }}}  e_{\cdot j}  \mathbf{\tilde{x}}^{p}_{j} \Bigg) \Big|   \\
& \leq \nu_{\ell}\Big|  \sum_{n=1}^{N}  \Bigg( \sigma^{'} \Big( \sum_{\mathclap{\substack{j\in \\ \Na(\mathbf{x})}}}  e_{\cdot j}  \mathbf{x}^{p}_{j}  \bm{\uptheta}_{p, S, t} \Big)  \sum_{\mathclap{\substack{j\in \\ \Na(\mathbf{x}) }}}  e_{\cdot j}  \mathbf{x}^{p}_{j} \Bigg) \Big|  +  \nu_{\ell}\Big|  \sum_{n=1}^{N}  \Bigg( \sigma^{'}\Big( \sum_{\mathclap{\substack{j\in \\ \Na(\mathbf{\tilde{x}} )}}}  e_{\cdot j}  \mathbf{\tilde{x}}^{p}_{j}   \bm{\uptheta}_{p, S^{i}, t} \Big)  \sum_{\mathclap{\substack{j\in \\ \Na(\mathbf{\tilde{x}} ) }}}  e_{\cdot j}  \mathbf{x}^{p}_{j} \Bigg) \Big|  \\
& \text{Using the fact that  the first order derivative is bounded,  } \\
& \leq 2\nu_{\ell} \alpha_{\sigma} N \mathbf{g}_{p, \lambda}  \\
\end{split}
\raisetag{6\normalbaselineskip}
\end{equation}
This completes the proof of Lemma~\ref{lemma:sgd_term2}. \\

\begin{lemma}[\textbf{Universal Graph Encoder    SGD Stability Bound}]\label{lemma:sgd_terms} \textit{ Let the loss \& activation functions be  Lipschitz-continuous and smooth. Let $\bm{\uptheta}_{S,T}$ and $\bm{\uptheta}_{S^i,T}$ denote the graph filter parameters of (single-layer)  Universal Graph Encode	 trained using SGD for $T$ iterations on two training datasets  $S$ and $S^{i}$, respectively. Then the expected difference in   the filter  parameters  is bounded by,}
	$$ \mathbf{E}_{\textsc{sgd}}  \big[\big| \bm{\uptheta}_{p, S,T} -\bm{\uptheta}_{p, S^i,T}  | \big]   \leq    \frac{2\eta \nu_{\ell} \alpha_{\sigma}  N\mathbf{g}_{p, \lambda}  }{m}  \sum_{t=1}^{T} \big( 1 + \eta\nu_{\ell}  \nu_{\sigma} N\mathbf{g}_{p, \lambda}^2 \big)^{t-1}  $$
\end{lemma}
\noindent \textbf{Proof}: Following the bounding analysis presented in~\cite{verma2019stability} for expected weight difference due to SGD, we have,

\begin{equation}~\label{eq:sgd_terms}
\begin{split}
& \mathbf{E}_{\textsc{sgd}}  \big[\big|\Delta \bm{\uptheta}_{p, t+1} | \big]   \leq \Big(1-\frac{1}{m}\Big) \mathbf{E}_{\textsc{sgd}}  \Big[ \Big| \Big(\bm{\uptheta}_{p, S, t}  - \eta  \nabla_{p} \ell\big(f(\mathbf{x},\bm{\uptheta}_{  S, t}),y \big) \Big) -  \Big(\bm{\uptheta}_{p, S^i, t}   - \eta  \nabla_{p} \ell\big(f(\mathbf{x},\bm{\uptheta}_{  S^{i}, t}), y \big) \Big) \Big| \Big]  \\ & \hspace{3em}  + \Big(\frac{1}{m}  \Big) \mathbf{E}_{\textsc{sgd}}  \Big[ \Big| \Big(\bm{\uptheta}_{p, S, t}    - \eta  \nabla_{p} \ell\big(f(\mathbf{\tilde{x}},\bm{\uptheta}_{  S,t}), \tilde{y} \big) \Big)  -   \Big(\bm{\uptheta}_{p, S^i, t}   - \eta  \nabla_{p} \ell\big(f(\mathbf{\hat{x}} ,\bm{\uptheta}_{S^{i}, t}), \hat{y} \big) \Big) \Big| \Big]  \\
& \leq \Big(1-\frac{1}{m}\Big) \mathbf{E}_{\textsc{sgd}}  \big[|\Delta \bm{\uptheta}_{p, t} | \big] + \Big(1-\frac{1}{m}\Big) \eta\mathbf{E}_{\textsc{sgd}}  \Big[ \Big|   \nabla_{p} \ell\big(f(\mathbf{x},\bm{\uptheta}_{ S, t}),y \big)   -    \nabla_{p}
\ell\big(f(\mathbf{x},\bm{\uptheta}_{  S^{i}, t}),y\big)  \Big| \Big]  + \\ & \hspace{1.5em} \Big(\frac{1}{m}\Big) \mathbf{E}_{\textsc{sgd}}  \big[|\Delta \bm{\uptheta}_{p, t} | \big]  +  \Big(\frac{1}{m}\Big) \eta\mathbf{E}_{\textsc{sgd}}  \Big[ \Big|   \nabla_{p} \ell\big(f(\mathbf{\tilde{x}} ,\bm{\uptheta}_{ S, t}), \tilde{y}  \big)   -  \nabla_{p} \ell\big(f(\mathbf{\hat{x}} ,\bm{\uptheta}_{ S^{i}, t}),\hat{y}  \big)  \Big| \Big]  \\ 
\end{split}
\end{equation}
\begin{equation*}
\begin{split}
& = \mathbf{E}_{\textsc{sgd}}  \big[|\Delta \bm{\uptheta}_{p, t} | \big] +   \Big(1-\frac{1}{m}\Big) \eta\mathbf{E}_{\textsc{sgd}}  \Big[ \Big|   \nabla_{p} \ell\big(f(\mathbf{x},\bm{\uptheta}_{  S, t}),y \big)   -      \nabla_{p} \ell\big(f(\mathbf{x},\bm{\uptheta}_{  S^{i}, t}),y\big)   \Big| \Big]   + \\ &    \hspace{1.5em}  \Big(\frac{1}{m}\Big) \eta\mathbf{E}_{\textsc{sgd}}  \Big[ \Big|   \nabla_{p} \ell\big(f(\mathbf{\tilde{x}} ,\bm{\uptheta}_{  S, t}), \tilde{y}  \big)   -   \nabla_{p} \ell\big(f(\mathbf{\hat{x}} ,\bm{\uptheta}_{  S^{i}, t}), \hat{y}  \big)  \Big| \Big]    \\ 
\end{split}
\end{equation*} \\

Plugging the bounds in	Lemma~\ref{lemma:sgd_term1} and Lemma~\ref{lemma:sgd_term2} into Equation~(\ref{eq:sgd_terms}),  we have,
\begin{equation*} 	
\begin{split}
& \mathbf{E}_{\textsc{sgd}}  \big[\big|\Delta \bm{\uptheta}_{p, t+1} | \big] \leq \mathbf{E}_{\textsc{sgd}}  \big[|\Delta \bm{\uptheta}_{p, t} | \big] + \Big(1-\frac{1}{m}\Big) \eta\nu_{\ell}  \nu_{\sigma}  N \mathbf{g}_{p, \lambda}^2    \mathbf{E}_{\textsc{sgd}}   [ |     \bm{\uptheta}_{p, t}|  ]   + \Big(\frac{1}{m}\Big) 2\eta \nu_{\ell} \alpha_{\sigma} N\mathbf{g}_{p, \lambda} \\ 
& = \Big( 1 + \big(1-\frac{1}{m}\big) \eta\nu_{\ell}  \nu_{\sigma} N \mathbf{g}_{p, \lambda}^2 \Big)   \mathbf{E}_{\textsc{sgd}}   [ |     \bm{\uptheta}_{p, t}|  ]    + \frac{2\eta \nu_{\ell} \alpha_{\sigma} N\mathbf{g}_{p, \lambda} }{m}  \\ 
& \leq \Big( 1 + \eta\nu_{\ell}  \nu_{\sigma} N\mathbf{g}_{p, \lambda}^2  \Big)   \mathbf{E}_{\textsc{sgd}}   [ |     \bm{\uptheta}_{p, t}|  ]    + \frac{2\eta \nu_{\ell} \alpha_{\sigma} N \mathbf{g}_{p, \lambda}   }{m}   \\
\end{split}
\end{equation*}

Lastly,   solving the $\mathbf{E}_{\textsc{sgd}} \big[\big|\Delta\bm{\uptheta}_{t} | \big]$ first order recursion yields, 
\begin{equation*} 	
\begin{split}	
& \mathbf{E}_{\textsc{sgd}}  \big[\big|\Delta \bm{\uptheta}_{p, T} | \big]   \leq    \frac{2\eta \nu_{\ell} \alpha_{\sigma} N \mathbf{g}_{p, \lambda} }{m}  \sum_{t=1}^{T} \big( 1 + \eta\nu_{\ell}  \nu_{\sigma} N \mathbf{g}_{p, \lambda}^2 \big)^{t-1}  \\
\end{split}
\end{equation*}
This completes the proof of Lemma~\ref{lemma:sgd_terms}. \\

\noindent \textbf{Bound on $\mathbf{g}_{p, \lambda}$: }   Let   $q=|\Na(\mathbf{x})|$ and $g_{\mathbf{x}}(\mathbf{L}) \in \mathbb{R}^{q \times q}$ be the submatrix of  $g(\mathbf{L})$ whose row and column indices are from the set $\{j \in  \Na(\mathbf{x})\}$.  We use $\mathbf{h}_{\mathbf{x, p}} \in \mathbb{R}^{q}$ to denote the $p^{th}$ moment graph signal  (node features) on the ego-graph $G_\mathbf{x, p}$.  Without loss of generality, we will assume that node $\mathbf{x}$ is represented by index $0$ in   $G_\mathbf{x, p}$. Thus, we can compute $ \sum_{j \in \Na(\mathbf{x})} e_{\cdot j}  \mathbf{x}^{p}_{j} = [g_{\mathbf{x}}(\mathbf{L}) \mathbf{h}_{\mathbf{x}, p}]_0$, a scalar value. Here $[\cdot]_0 \in \mathbb{R}$ represents the value of a vector at index 0, i.e., corresponding to  node $\mathbf{x}$. Then the following holds (assuming the graph signals are normalized, i.e., $\| \mathbf{h}_{\mathbf{x}, 1} \|_{2}=1$),
\begin{equation}~\label{eq:g_lambda} 	
\begin{split}
|[g_{\mathbf{x}}(\mathbf{L}) \mathbf{h}_{\mathbf{x}, p}]_0| \leq \|g_{\mathbf{x}}(\mathbf{L})  \mathbf{h}_{\mathbf{x}, p} \|_{1} \leq \|g_{\mathbf{x}}(\mathbf{L})\|_2  \|\mathbf{h}_{\mathbf{x}, p} \|_{2}  \leq  \|g_{\mathbf{x}}(\mathbf{L})\|_2  \|\mathbf{h}_{\mathbf{x}, 1} \|_{2} =   \lambda_{G_\mathbf{x}}^{\max}   
\end{split} 
\end{equation}
where the third inequality follows from the fact that $\sum_{i} |x^{p}_{i}| \leq \sum_{i} |x_{i}|^{2} \leq 1 $ for $p \geq 2$ and since $  \|x\|_2 = 1$. As a result the norm inequality $(\sum_{i} |x^{p}_{i}|)^{1/2} \leq   (\sum_{i} |x_{i}|^{2})^{1/2} \leq 1 $ holds.

Finally, plugging $\mathbf{g}_{p, \lambda} \leq \lambda_{G}^{\max}$ and Lemma~\ref{lemma:sgd_terms} into Equation~(\ref{eq:proof_part1}) yields the following remaining result,

\begin{equation*} 	
\begin{split}
& 2\beta_{m}  \leq   \sum_{p=1}^{P}  \Big(  \alpha_{\ell} N \lambda_{G}^{\max}   \mathbf{E}_{\textsc{sgd}}\big[\big|\Delta \bm{\uptheta}_{p} \big|\big]  \Big)    \\      
& \beta_{m}  \leq   \frac{ P \eta \alpha_{\ell}   \alpha_{\sigma} \nu_{\ell} N^2 (\lambda_{G}^{\max})^2 \sum_{t=1}^{T} \big( 1 + \eta\nu_{\ell}  \nu_{\sigma} N (\lambda_{G}^{\max})^2 \big)^{t-1}}{m}    \\  
&   \beta_{m}  \leq  \frac{1}{m}\BigO \Big(P N^{T+1} (\lambda_{G}^{\max})^{2T}\Big)  \hspace{2em} \forall T \geq 1  \\ 
\end{split} 
\end{equation*}

This completes the full proof of Theorem 3.

\section{Significance of Adaptive Supervised Graph Kernel Loss }

\begin{figure}[h!]
	
	\begin{subfigure}[t]{0.45\textwidth}
		\centering
		\includegraphics[width=1.1\textwidth]{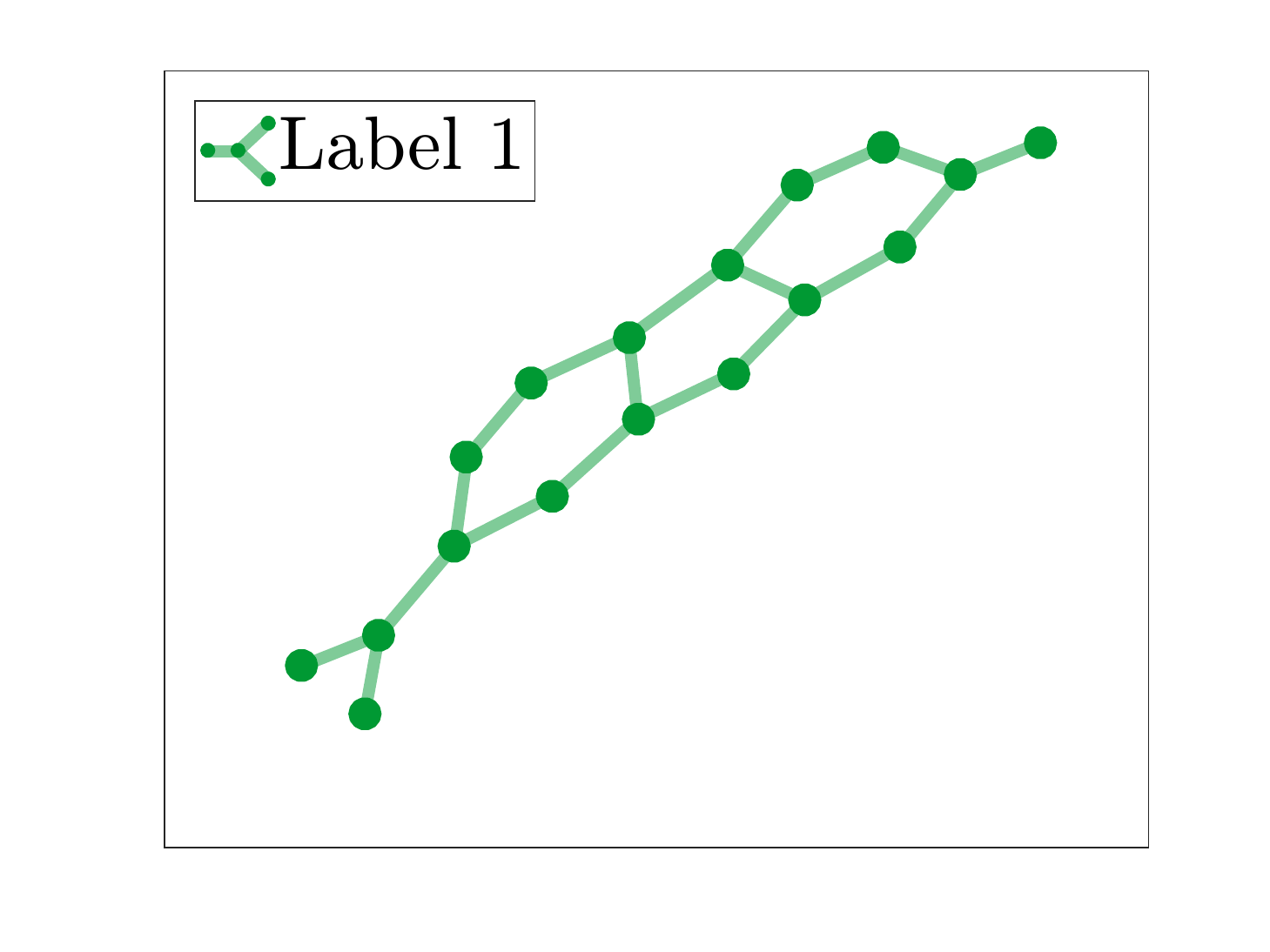}
		\vspace{-2em}
		\caption{Typical label $1$ sample.}
		\label{fig:l1_sample}
	\end{subfigure}
	\begin{subfigure}[t]{0.45\textwidth}
		\centering
		\includegraphics[width=1.1\textwidth]{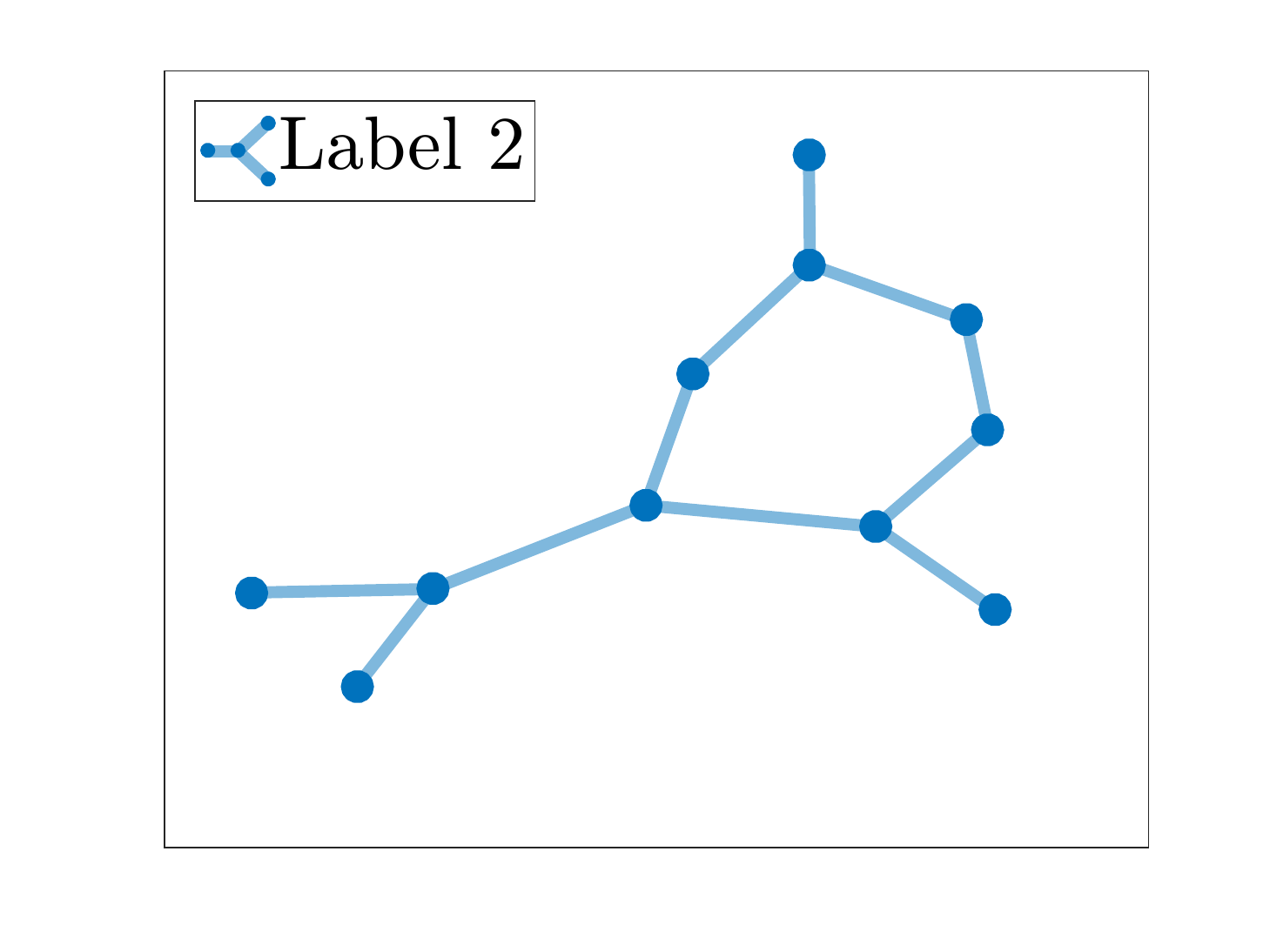}
		\vspace{-2em}
		\caption[short]{Typical label $2$  sample. }
		
		\label{fig:l2_sample}
	\end{subfigure} 
	
	\caption{Some MUTAG graph data samples.}\label{fig:projection-results}
	
\end{figure}

We qualitatively demonstrate the    significance of employing the adaptive supervised graph kernel loss on MUTAG dataset. For this purpose, we take a deeper dive into the results to find out which sub-structures are more important for prediction.  Figures~\ref{fig:l1_sample} and \ref{fig:l2_sample} depict some representative  graph structures for class labels $1$ and $2$   on the MUTAG dataset. It turns out that the graph samples with label $1$ have mostly three or more cycles, while the graph samples with  label $2$ tend to contain only one cycle. On the other hand, the graph samples with two cycles can belong to either class. By  simply  creating a learning rule that data samples containing $3$ or more cycles belong to label $1$, we can get a prediction  accuracy around $84\%$. This simple rule alone beats the random-walk graph kernel based method, which achieves a prediction accuracy of $80.72\%$~\cite{shervashidze2011weisfeiler}. By employing an adaptive supervised graph kernel, the model can thus learn  embeddings which are biased more towards a graph kernel that better captures the count of number of cycles in graphs, and discount the graph kernels which attempt to match the random walk distributions which do not help with increasing the prediction accuracy.

\section{Experiment Baselines Settings }

For the Weisfeiler-Lehman (WL) kernel, we vary the iteration pararmeter $h\in\{2, 3, 4, 5\}$. For the Random-Walk (RW) kernel, the decay factor is chosen from $\{10^{-6},10^{-5}...,10^{-1}\}$. For the graphlet kernel (GK), we choose the graphlet  size $\{3,5,7\}$.  For the deep graph kernels (DGKs),  the  window size and dimension are taken from the set $\{2,5,10,25,50\}$ and report the best classification accuracy obtained among i) deep graphlet kernel, ii) deep shortest path kernel and iii) deep Weisfeiler-Lehman kernel. For the Multiscale Laplacian Graph (MLG) kernel, we vary the $\eta$ and $\gamma$ parameters of the algorithm from $\{0.01,0.1,1\}$, radius size from $\{1,2,3,4\}$, and level number from $\{1,2,3,4\}$. For  the diffusion-convolutional neural networks (DCNN), we choose the number of hops from $\{2,5\}$ and employ the AdaGrad algorithm (gradient descent) with the following parameters: the learning rate $0.05$, batch size $100$ and number of  epochs $500$. We use the node degree   as the labels  in the cases where node labels are unavailable.  For the rest, best reported results are borrowed from their respective papers since the experimental  setup is same and fair comparison can be made.

\section{Ablation Studies and Discussion}\label{sec:ablation_study}

\noindent \textbf{How   effective is adaptive supervised graph kernel loss?}

\renewcommand{\arraystretch}{2}
\begin{SCtable}[\sidecaptionrelwidth][h!]
	\centering
	\fontsize{7}{8}\selectfont
	\begin{tabular}{ p{3.0cm} |     K{1.6cm}  !{\vrule width0.8pt} K{1.6cm} !{\vrule width0.8pt} K{1.6cm}   | }
		
		\hspace{-0.8em}	\multirow{1}{*}{\textbf{Model / Dataset}} &       	\multicolumn{1}{c!{\vrule width0.8pt}}{PTC}  &  \multicolumn{1}{c!{\vrule width0.8pt}}{ENZYMES}      \\ \hline
		
		
		\hspace{-0.8em}	\textsc{DUGnn}  &     $73.53$ 	    &  $65.00$    \\  \hline
		\rowcolor[RGB]{252, 247, 246}  
		\hspace{-0.8em}	\textsc{DUGnn} - $\La_{\Ka}^{(\text{unsup})}$ + $\La_{\Ka}^{(\text{sup})}$   &    $76.47$  	    &   $64.13$   \\  \hline

		\Xhline{2\arrayrulewidth}
	\end{tabular}	
	\caption{Ablation Study of Supervised Adpative Graph Kernel Loss. \textsc{Dugnn} is the base model trained with non-adaptive kernel loss function. 	\textsc{DUGnn}   - $\La_{\Ka}^{(\text{unsup})}$ + $\La_{\Ka}^{(\text{sup})}$  is trained with adaptive  loss inplace of non-adaptive graph kernel  loss. } 
	\label{table:adpative_perform}
	
\end{SCtable}

We observe that employing the adaptive supervised graph kernel loss yields a smoother decay in the validation loss and produces more stable results. However  as evident from Table~\ref{table:adpative_perform}, it only increases the performance  on PTC by $\mathbf{3\textbf{\%}}$,  and  reducess the performance on ENZYMES by around $\mathbf{1\textbf{\%}}$. As a result, we advice   treating the adaptive supervised kernel loss as a hyper-parameter.

\end{document}